\documentclass[journal]{IEEEtran}
\usepackage[font=small,labelfont=bf,justification=justified]{caption}
\usepackage{cite}

\ifCLASSINFOpdf
  \usepackage[pdftex]{graphicx}
\else
  \usepackage[dvips]{graphicx}
\fi
\usepackage{amsmath}
\ifCLASSOPTIONcompsoc
  \usepackage[caption=false,font=normalsize,labelfont=sf,textfont=sf]{subfig}
\else
 \usepackage[caption=false,font=footnotesize]{subfig}
\fi

\ifCLASSOPTIONcaptionsoff
  \usepackage[nomarkers]{endfloat}
 \let\MYoriglatexcaption\caption
 \renewcommand{\caption}[2][\relax]{\MYoriglatexcaption[#2]{#2}}
\fi
\usepackage{url}

\usepackage{multirow}
\usepackage{ctable}
\DeclareMathOperator*{\argmax}{arg\,max}

\begin{document}
%
\title{Generating Hard Examples for Pixel-wise Classification}
%

\author{Hyungtae~Lee,~\IEEEmembership{Member,~IEEE,}
        Heesung~Kwon,~\IEEEmembership{Senior Member,~IEEE}
        and~Wonkook~Kim
\thanks{Manuscript received XX, 2018;}
\thanks{Hyungtae Lee and Heesung Kwon are with the Intelligent perception branch, the Computational and Information Sciences Division (CISD), Army Research Laboratory, Adalphi, MD, USA, 20783 (e-mail: \{hyungtae.lee,heesung.kwon\}.civ@mail.mil).}
\thanks{Wonkook Kim is with Department of Civil and Environmental Engineering, Pusan National University, Busan, Korea, 46241 (e-mail: wonkook@pusan.ac.kr).}
\thanks{© 2022 IEEE. Personal use of this material is permitted. Permission from IEEE must be obtained for all other uses, in any current or future media, including reprinting/republishing this material for advertising or promotional purposes, creating new collective works, for resale or redistribution to servers or lists, or reuse of any copyrighted component of this work in other works.}}

%
%

\markboth{IEEE Journal of Selected Topics in Applied Earth Observations and Remote Sensing}
{Shell \MakeLowercase{\textit{et al.}}: Bare Demo of IEEEtran.cls for IEEE Journals}
%



\maketitle

\begin{abstract}

Pixel-wise classification in remote sensing identifies entities in large-scale satellite-based images at the pixel level. Few fully annotated large-scale datasets for pixel-wise classification exist due to the challenges of annotating individual pixels. Training data scarcity inevitably ensues from the annotation challenge, leading to overfitting classifiers and degraded classification performance. The lack of annotated pixels also necessarily results in few hard examples of various entities critical for generating a robust classification hyperplane. To overcome the problem of the data scarcity and lack of hard examples in training, we introduce a two-step hard example generation (HEG) approach that first generates hard example candidates and then mines actual hard examples. In the first step, a generator that creates hard example candidates is learned via the adversarial learning framework by fooling a discriminator and a pixel-wise classification model at the same time. In the second step, mining is performed to build a fixed number of hard examples from a large pool of real and artificially generated examples. To evaluate the effectiveness of the proposed HEG approach, we design a 9-layer fully convolutional network suitable for pixel-wise classification. Experiments show that using generated hard examples from the proposed HEG approach improves the pixel-wise classification model’s accuracy on red tide detection and hyperspectral image classification tasks. 

\end{abstract}


%
\IEEEpeerreviewmaketitle

\section{Introduction}
\label{sec:intro}

\IEEEPARstart{P}{ixel-wise} classification is the task of identifying entities at the pixel level in remotely sensed images, such as Earth-observing satellite-based images from multi- or hyperspectral imaging sensors. The pixel-wise classification has some parallels to image segmentation. Still, there are several limitations to directly using the state-of-the-art image segmentation methods for the pixel-wise classification. Image segmentation methods treat an image as a composition of multiple instances of a scene or object and delineate boundaries between different instances. Current state-of-the-art image segmentation methods adopt the ability to segment these instances either by using a joint detection and segmentation model~\cite{KHeICCV2017} or by finetuning a detection model~\cite{JLongCVPR2015}. However, these detection abilities are only useful if the target object or scene provides category-specific contextual or structural information and if each instance covers a relatively large area of the image. Unfortunately, these requirements are not typically met in remotely sensed images; thus, spectral characteristics embedded in each pixel are used as viable pixel-wise classification information.

\begin{figure}[t]
\captionsetup{font=small}
\centering
\centerline{\includegraphics[width=\linewidth,trim=5mm 5mm 5mm 5mm,clip]{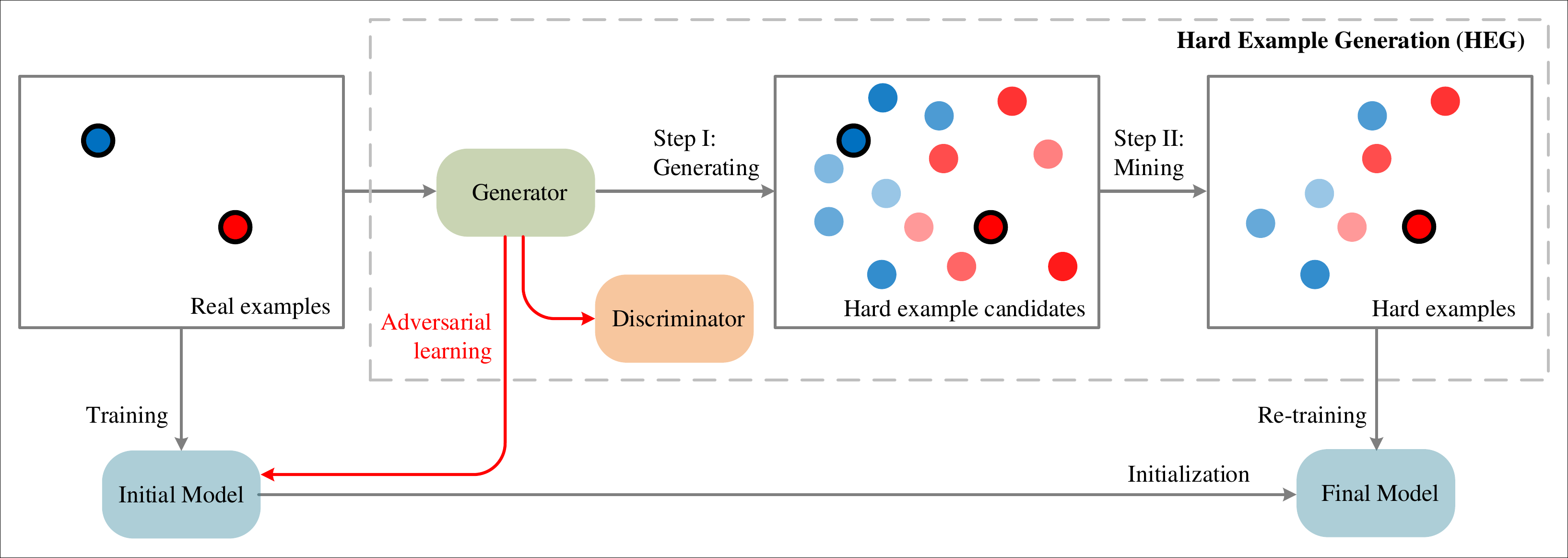}}
\caption{{\bf Hard Example Generation (HEG) approach.} While the initial model for pixel-wise classification is trained using real examples, the final model is initialized from this initial model and re-trained with hard examples that are the output of the HEG approach. The HEG approach takes two steps: i) generation of hard example candidates and ii) hard example mining. In the first step, the generator is trained via the adversarial learning framework to fool the initial pixel-wise classification and the discriminator that distinguishes real examples from generated examples. For the red tide detection task, we created hard negatives by applying the generator only to the negative examples, as we aim to solve the problem of the lack of hard negative examples. On the other hand, for the hyperspectral image classification task, hard examples are generated for all categories as shown in this figure.}
\label{fig:heg_process}
\end{figure}

One of the issues with pixel-wise classification for remote sensing images is the lack of fully annotated large-scale remote sensing datasets. Since it is exceptionally challenging to annotate each pixel of the remote sensing image, frequently, many pixels in the image remain unlabeled, leading to performance degradation. Further performance decrease is caused by sparse training data, including few hard examples necessary to generate a robust classification hyperplane.

To address the lack of hard examples, we introduce a hard example generation approach (HEG) suitable for pixel-wise classification (Figure~\ref{fig:heg_process}). The proposed HEG approach takes two steps: i) generating hard example candidates that were recognized as false positives for other categories while preserving the properties of the original category (generation step) and ii) processing hard example mining to discover hard examples incorrectly detected with high loss (mining step).

In the first step, we use a variant of the generative adversarial learning (GAN)~\cite{IGoodfellowNIPS2014} to train a generator that creates hard example candidates. To prevent the generated examples from losing the specific property of its corresponding category, we trained a network to distinguish the real examples from the artificially generated examples, which serves as \emph{discriminator} in the GAN framework. In order for the generated example to be a hard example for another category while preserving the original category's properties, a pixel-wise classification model and the discriminator become the counterparts that the generator should deceive.

For the second step, we redesigned the online hard example mining (OHEM)~\cite{AShrivastavaCVPR2016} to select hard examples. Shrivastava et al. introduced OHEM, which, for every iteration, sampled a small number of examples with a high loss from the overwhelming examples and use them in training CNN. We used a simple variant of OHEM to mine a huge volume of artificially generated examples in a cascaded way. First, we randomly choose a subset of examples that can be either real or artificially generated examples and then apply OHEM to the selected examples to choose hard examples.

\begin{figure*}[t]
\captionsetup{font=small}
\centering
\centerline{\includegraphics[width=\linewidth,trim=5mm 5mm 5mm 5mm,clip]{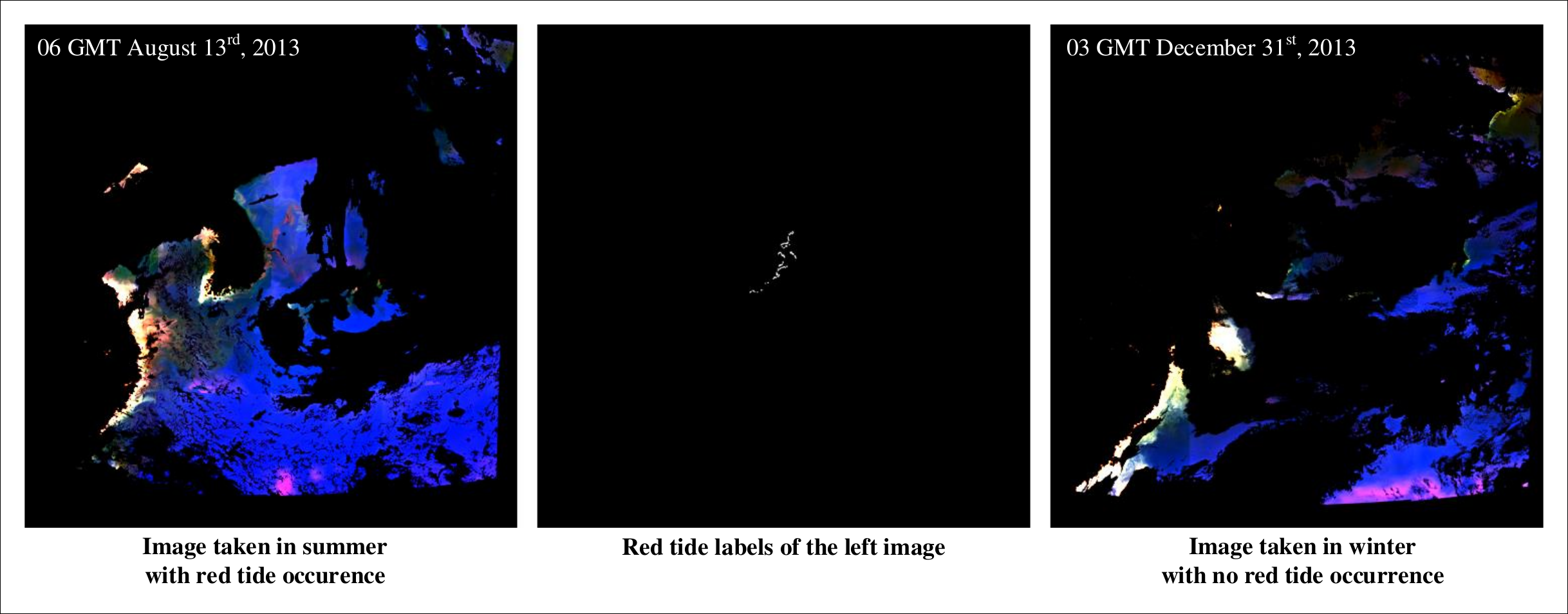}}
\caption{{\bf GOCI Satellite Images and Red Tide Labels.} Summer (e.g., an image taken on August) and winter (e.g., an image taken on December) images of the same area are placed on the left and right of the figure, respectively. In the middle figure, the red tide labels of the summer image are marked as white pixels. In the summer image, not all red tide regions are labeled, and not all the regions in red represent red tide. All the red regions that appear in winter images are not red tide regions.}
\label{fig:goci_img}
\end{figure*}

To evaluate the proposed HEG approach, we implemented a 9-layer fully convolutional network (FCN) inspired by~\cite{HLeeTIP2017}. The FCN architecture has proved to be suitable for pixel-wise classification~\cite{HLeeIGARSS2018,HLeeIGARSS2016,HLeeTIP2017,HLeeIGARSS2019}. We validate our approach to red tide detection using the large-scale remote sensing image dataset obtained from multi-spectral GOCI (Geostationary Ocean Color Imager)~\cite{SChoKJRS2010} on a geostationary satellite. We chose this practical task because it clearly presents the ground-truthing problems mentioned earlier\footnote{Since the biological properties of red tide are not clearly visible in the image, we used the information on real-world red tide occurrences reported by NIFS (National Institute of Fisheries Science) (\url{http://www.nifs.go.kr/red/main.red/}) of South Korea. NIFS manually examined red tide occurrence only at a limited number of locations along the southern seashore of South Korea.}. Due to such challenging ground-truthing problems inherent in remote sensing, red tide occurrences are labeled only at a limited number of locations. Moreover, there are no labels about where no red tide was found that could be used as negative examples in training. Therefore, we end up with only a small number of spectral examples from a fraction of areas where red tide occur in training. In this work, we use the images taken in December as negative examples where red tides do not occur due to the low water temperature\footnote{In South Korea, summer is in July and August and winter in December. Red tide occurs mainly in summer when the water temperature is high.}. Figure~\ref{fig:goci_img} shows the GOCI images used for the positive (red tide) and negative (non-red tide) training examples, and the red tide region annotation of the positive image.

From this peculiar GOCI image setting, we found severe issues highlighting the need for the proposed HEG approach. First, the spectral characteristics of the images taken in December are very different from those of the images taken in the summer when red tides mostly occur because the marine environment in summer and winter is very different. Therefore, the negative examples from the December images do not generally represent the non-red tide area from the images collected in the summer. Second, the imbalance between the numbers of positive examples and the negative examples is quite significant. The number of positive examples is in the order of one hundred pixels per image. In comparison, the number of negative examples is about 31M pixels per image as all the pixels of the GOCI image (5567$\times$5685) taken in winter are used as negative examples. Lack of non-red tide examples derived from the property discrepancy between the summer and winter images associated with the first problem is addressed by the first step of HEG (i.e., generation of hard example candidates). The second problem of the data imbalance between the positive and negative examples is effectively alleviated by the second step of HEG (i.e., hard example mining).

We conducted extensive experiments to determine how the proposed HEG addresses the problems that arise in training the pixel-wise classification model on GOCI images. For red tide detection, a one-class classification problem with significantly unbalanced distribution, we use HEG to generate hard negative examples. To show that the proposed HEG can be easily extended to other tasks with multiple categories, we also applied it to several pixel-wise hyperspectral image classification tasks. Experiments have confirmed that pixel-wise classification method trained by adopting the proposed HEG significantly enhances performance for red tide detection and several hyperspectral image classification tasks.

\section{Related Works}
\label{sec:related_work}

\noindent{\bf Training generator via adversarial learning.} Szegedy et al.~\cite{CSzegedyICLR2014} introduced a method to generate an adversarial image by adding perturbation to be misclassified by a CNN-based recognition approach. These perturbed images become adversarial images to the recognition approach. Goodfellow et al.~\cite{IGoodfellowNIPS2014} introduce two models: a generator that captures the data distribution and a discriminator that estimates the probability that an example came from the training data rather than the generator. A generator and a discriminator are trained at the same time in a direction to interfere with each other. This is called an adversarial learning framework. 

Radford et al.~\cite{ARadfordICLR2016} devised an image generation approach based on CNN by adopting this adversarial learning framework. Wang et al.~\cite{XWangCVPR2017} used the adversarial learning framework to train a network that creates artificial occlusion and deformation on images. An object detection model is trained against this adversary to improve performance. Hughes et al.~\cite{LHHughesRS2018} introduce a negative generator based on an autoencoder that takes a positive image as an input. The generator is optimized to make the output acquire the properties of a real image by adopting an adversarial learning framework. This generator is used to augment the set of negative examples, which are not necessarily hard negatives. Choi et al.~\cite{JChoiICCV2019} use GAN-based data augmentation for reducing a domain gap from fully annotated synthetic data to unsupervised data. Xie et al.~\cite{CXieCVPR2020} also uses adversarial learning to augment training examples for image recognition. In the proposed work, we also use adversarial learning to train a hard example generator (HEG). Unlike~\cite{LHHughesRS2018,CXieCVPR2020,JChoiICCV2019}, our HEG generates hard negative examples, which are more challenging to be identified as non-red-tide examples by our red tide detector than other real negative examples.\medskip

\noindent{\bf Hard example mining.} Sung and Poggio~\cite{KSungMITAIMemo1994} first introduced hard negative mining (also known as bootstrapping) that trains the initial model with randomly chosen negatives and adapts the model to hard negatives that consist of false positives of the initial model. Thereafter, hard example mining has been widely used in various applications such as pedestrian detection~\cite{LBourdevECCV2010,NDalalCVPR2005}, human pose estimation~\cite{LBourdevICCV2009,HLeeACCV2012}, action recognition~\cite{SMajiCVPR2011,HLeeCVPRW2014}, event recognition~\cite{HLeeWACV2015}, object detection~\cite{PFelzenszwalbTPAMI2010,RGirshickTPAMI2016,KHeTPAMI2015,TMalisiewiczICCV2011}, and so on. There are alternative ways to find hard examples using heuristic~\cite{RGirshickICCV2015} or other hard example selection algorithms~\cite{SRenTPAMI2017,JUijlingsIJCV2013}, which avoid training multiple times. Kellenberger et al.~\cite{BKellenbergerTGARS2019} use active learning, which requires human intervention to assign labels to critical examples and address the problem of the small number of positive samples.

Shrivastava et al.~\cite{AShrivastavaCVPR2016} introduced online hard example mining, which, for every training iteration, carries out hard example mining that chooses examples with high training loss. However, it is too exhaustive to evaluate all examples on each iteration. Hence, when using an extensive example set like our problem, it is impractical to examine all examples for each iteration. Therefore, we use OHEM in a cascaded fashion to randomly select a subset of examples and then perform efficient mining on it.\medskip

\noindent{\bf CNN used for detecting natural phenomena in marine environment.} Since CNN has provided promising performance in image classification~\cite{GChenTGARS2018,XFengTGARS2020,XYaoTGARS2021}, there have been several attempts to use it in the marine environment. CNNs have been effectively used for detection of coral reefs~\cite{AKingCVPRW2018,AMahmoodICIP2016}, classification of fish~\cite{GDingOCEANS2017,ZGeICIP2015,RMandalIJCNN2018}, detection of oil from shipwreck~\cite{MNietoHidalgoTGARS2018,JZiniTGARS2020}, and so on. However, applying deep neural network to detect objects-of-interest in the marine environment has been quite limited due mainly to difficulties in acquiring large amounts of annotated data, unlike general object detection applications. In this paper, we devise a CNN training strategy coupled with an advanced network architecture tailored to red tide detection while minimizing human labeling efforts.

\section{Red Tide Detection Approach}
\label{sec:pixelCNN}

\subsection{Red Tide Detection}
\label{ssec:detection}

In this section, we describe the proposed CNN-based red tide detection approach. This approach takes the GOCI image as input and evaluates whether each pixel in the image belongs to a red tide area or not. Therefore, red tide detection can be considered as pixel-wise classification. The architecture of the proposed approach is built on a model introduced by~\cite{HLeeTIP2017}, which is known to be suitable for pixel-wise classification. We apply a sliding window method to deal with limited GPU memory when processing GOCI images.\medskip

\begin{figure*}[t]
\captionsetup{font=small}
\centering
\begin{minipage}[b]{\linewidth}
  \centerline{\includegraphics[width=\linewidth,trim=5mm 5mm 5mm 5mm,clip]{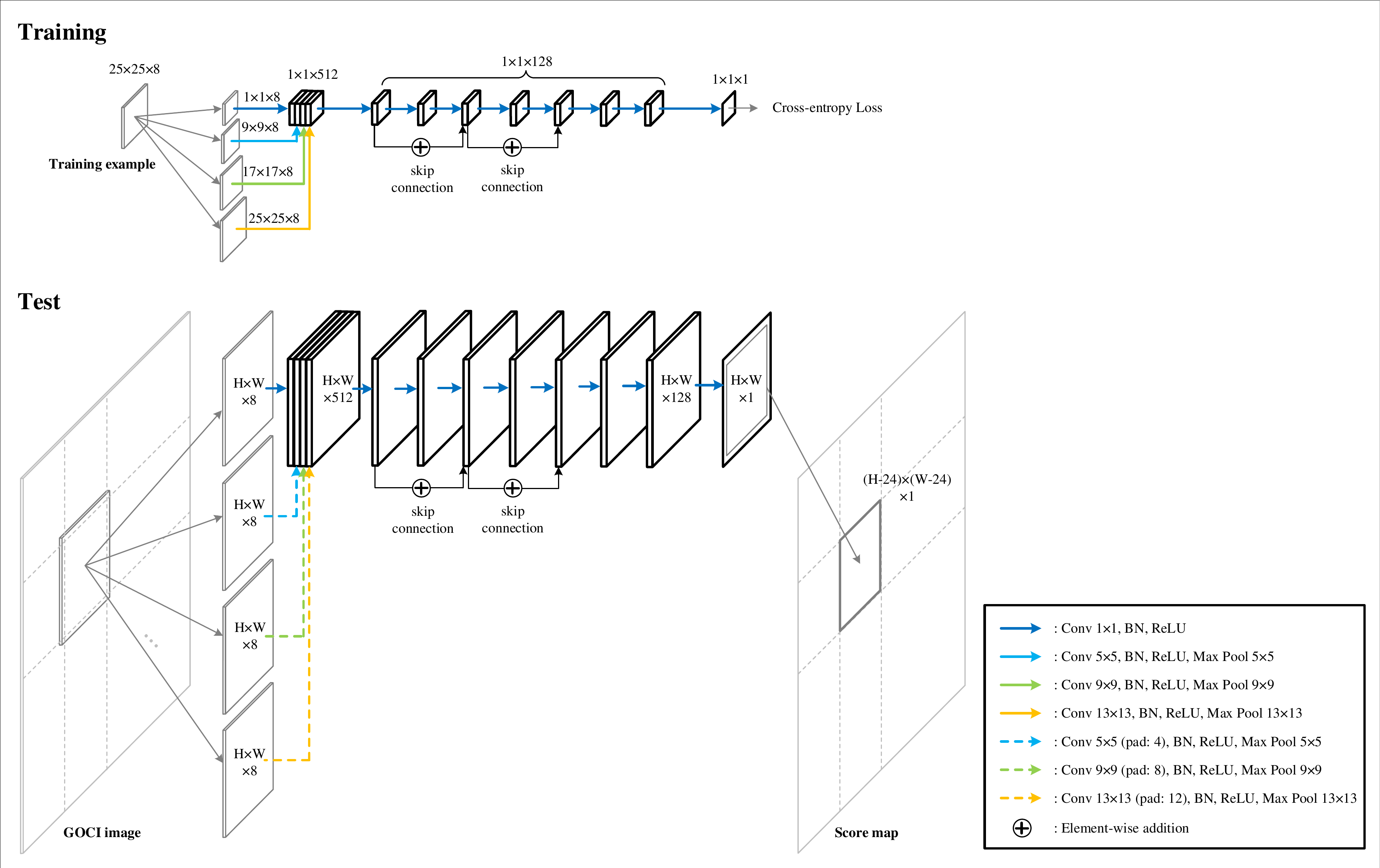}}
\end{minipage}
\caption{{\bf FCN Architecture for Red Tide Detection.} The training and test architectures have slightly different initial multi-scale filter banks tailored to pixel-wise classification. The weights and dimension of the convolutional layers do not change between training and test. The size of each intermediate feature blob is shown above the blob.
}
\label{fig:architecture}
\end{figure*}

\noindent{\bf Pixel-wise classification.} Pixel-wise classification has been widely used for mulitispectral/hyperspectral image classification that assigns each pixel vector into a corresponding category by exploiting the spectral characteristics of both the pixel and the neighboring pixels in a local region. Unlike general image segmentation~\cite{JLongCVPR2015,KHeICCV2017}, which segments distinctive scene components in an image by primarily leveraging object appearance as well as structural characteristics and attributes of the components (e.g., human anatomy, car with four wheels), pixel-wise classification is a task of predicting each pixel in a region with additional features, such as spectral profiles, and simultaneously little structural attributes available. Therefore, in the proposed work, red tide detection is treated as a pixel-wise classification problem primarily because the red tide is a microscopic alga with no discernible appearances or structures useful for image segmentation.

For recent CNN-based image segmentation, the state-of-the-art approaches have a CNN architecture designed as sequentially stacking multiple layers consisting of filters that capture neighboring information (e.g., 3$\times$3, 5$\times$5 filters) to leverage information over a large area when predicting each pixel. Furthermore, it adopts multiple downsampling layers such as pooling/convolution layers with stride$\geq$2, which are known to encode the structural characteristics adequately. On the other hand, our approach, described in the next section, also adopts CNN architecture. It is designed by stacking layers made up of 1$\times$1 filters except for the initial layers and does not use any downsampling layers. The first layer, consisting of multi-scale filters, does not capture structural characteristics but rather imposes spatial continuity over neighboring pixels such that they have the same identity.\medskip

\noindent{\bf Architecture.} The architecture of the proposed red tide detection model is shown in Figure~\ref{fig:architecture}. To cope with the pixel-wise classification of red tide detection, we use a 9-layer fully convolutional network (FCN), which intakes an image of arbitrary size. The network takes image patches of 25$\times$25 as input during training, while an image patch of a certain dimension determined by the maximum size of GPU memory is fed into the network in test time.

\begin{figure*}[t]
\captionsetup{font=small}
  \centering
\begin{minipage}[b]{\linewidth}
  \centerline{\includegraphics[width=\linewidth,trim=10mm 5mm 8mm 5mm,clip]{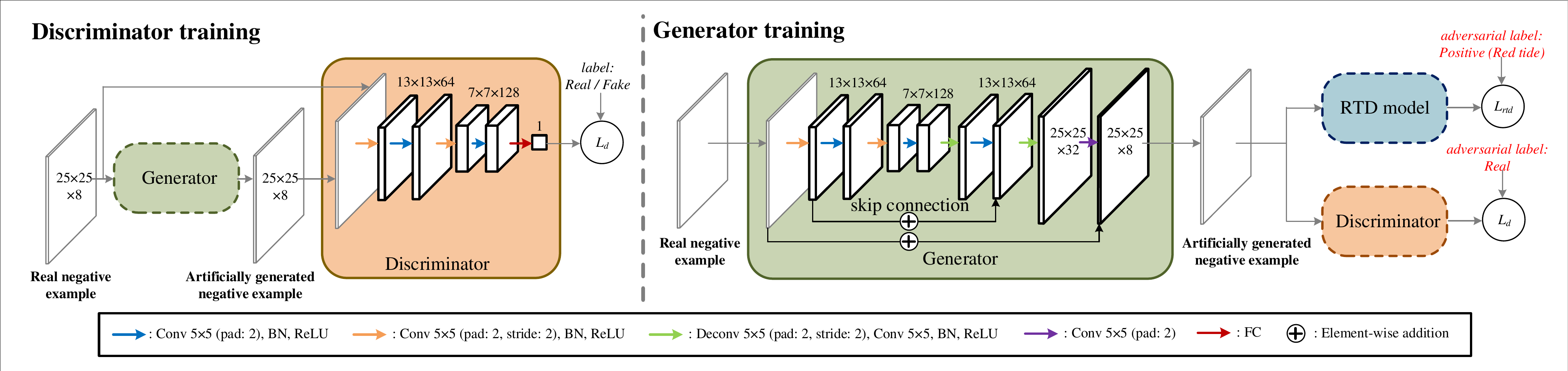}}
\end{minipage}
\caption{{\bf Training hard example candidates generator.} For each iteration in training, the discriminator and the generator are trained in order. Components not updated during training is indicated with a dashed box.
}
\label{fig:hng}
\end{figure*}

The network's initial module is a multi-scale filter bank consisting of convolutional filters with four different sizes (1$\times$1, 5$\times$5, 9$\times$9, and 13$\times$13). The architecture of the multi-scale filter bank is slightly different between training and testing. Given an image patch of 25$\times$25 in training, each $k\times k$ filter is convolved with a patch of $(2k-1)\times(2k-1)$ centered on the 25$\times$25 patch. The size of the smaller patch, i.e., $(2k-1)\times(2k-1)$, is determined so that each convolution always includes the center of the larger 25x25 patch. For example, when applying a 5$\times$5 filter to a 9$\times$9 patch, the 5$\times$5 window always contains the center pixel of the 25$\times$25 patch being evaluated. After the initial convolution, a max pooling is applied to the outputs of the convolutional filters so that those pooled feature maps have a size of 1$\times$1 except for the 1$\times$1 convolution. In the test, four filters are applied to the same patch of the same size. These convolutions use appropriate padding to ensure that the four pooled feature maps have the same size. Four output feature maps are concatenated for both training and testing and then fed to the second convolutional layer. Accordingly, due to the multi-scale filter bank architecture, our network becomes 25$\times$25, and the network’s receptive field uses spatial information based on this receptive field when evaluating each pixel.

In training, dropout layers, which are commonly used to solve the overfitting issue to some extent, are added at the end of the $7^{th}$ and $8^{th}$ layers. The rest of the network is the same in training and testing. Specifically, the binary sigmoid classifier that is useful for either single-label or multi-label classification is used for the output layer to identify other natural phenomena (e.g. sea fog, yellow dust, etc.) from the GOCI images later using the same architecture.\medskip

\noindent{\bf Sliding window strategy.} The proposed method cannot process a huge GOCI image at once during inference because of GPU memory limitations, where the size of the image is 5567$\times$5685. To overcome this issue, we use a sliding-window-based strategy, where each window size is $H\times W$. In our experiment, we set $H$ and $W$ to 600. Considering our network's receptive field, we use only the scores corresponding to the central $(H-24)\times(W-24)$ region as the final output. 

\subsection{Training: Adopting Hard Example Generation}
\label{ssec:training}

To meet the need for hard examples in devising an accurate hyperplane with a small number of examples that can be adequately applied to test examples, we introduce hard example generation (HEG) approach. It takes two steps: i) generation of hard example candidates (\emph{generation} step) and ii) hard example mining (\emph{mining} step). This section provides details for each step and our training strategy to jointly train the red tide detection model with HEG. Note that the details given are primarily focused on the red tide detection task but can easily be extended to other pixel-based classifications.\medskip

\noindent{\bf Generation step.} This step develops a generator that creates artificial examples that the red tide detection model likely classifies as false positives. The proposed generator creates hard negative example candidates only for single-category red tide detection. However, it can still be applied to other multi-category pixel-based classification tasks for generating hard example candidates for multi-category positives and negatives. This extension will be presented with hyperspectral image classification in Section~\ref{sec:hyperspectral}. The generator is designed as a 10-layered conv-deconv network consisting of eight convolutional layers and two deconvolutional layers inspired by U-Net with high image generation capability~\cite{HCaiCVPR2018,DFourureBMVC2017,WLiuCVPR2018,HNohICCV2015,ORonnebergerMICCAI2015}, as shown in Figure~\ref{fig:hng}.

We aim to achieve two goals in the training of the generator. First, the generator must be able to fool the red tide detection model so that the generated examples are incorrectly classified as red tides. Second, generated examples should have typical non-red-tide spectral characteristics. To achieve the goals, we introduce a discriminator that distinguishes real examples from artificially generated ones. The generator is trained to deceive the discriminator as in the typical GAN framework~\cite{IGoodfellowNIPS2014}. The discriminator consists of four convolutional layers and one fully-connected layer, as shown in Figure~\ref{fig:hng}. Generated examples become hard example candidates that are designed to maximize the losses of the red tide detection model and the discriminator, which conflicts with the two models' objectives. The training process of the generator and the discriminator is shown in Figure~\ref{fig:hng}.

To mathematically formulate the process of generating hard negative examples, the red tide detection model and its loss are represented by $\mathcal{F}_{rtd}$ and $\mathcal{L}_{rtd}$, respectively. The red tide detection model is trained by minimizing its loss expressed as:
\begin{equation}
    \mathcal{L}_{rtd}(E,L_{rtd}) = \mathcal{H}(\mathcal{F}_{rtd}{(E)},L_{rtd}),
\end{equation}
where $E$ and $L_{rtd}$ are training examples and their associated labels, respectively. For each example $e\in E$, its red tide labels $l_{rtd} \in L_{rtd}$ can be either 1 (red tide) or 0 (non-red tide). $\mathcal{H}(p,q)$ is the cross-entropy for the distributions $p$ and $q$.

The discriminator and its loss are denoted as $\mathcal{F}_{d}$ and $\mathcal{L}_{d}$, respectively, in Equation~\ref{eq:discr_train}. The discriminator is optimized by minimizing its loss which is expressed as:
\begin{equation}
    \mathcal{L}_{d}(N) = \mathcal{H}{(\mathcal{F}_{d}{(N)},\bf{1})} + \mathcal{H}{(\mathcal{F}_{d}{(\mathcal{F}_{g}{(N)})},\bf{0})},
    \label{eq:discr_train}
\end{equation}
where $N$ is a set of real negative examples. The discriminator's labels can be either 1 (real example) or 0 (artificially generated example). $\mathcal{F}_{g}$ denotes the generator. The randomly selected 256 negative examples and the generated examples are used for every iteration when training the discriminator. The generated examples are the output of the generator taking the randomly selected real negative examples as input.

The generator's objective is to generate negative examples incorrectly classified as red tide by the red tide detection model and as real negative examples by the discriminator. Accordingly, the generator loss ($\mathcal{L}_{g}$) can be expressed as:
\begin{equation}
    \mathcal{L}_{g}(N) = \mathcal{H}(\mathcal{F}_{rtd}{(\mathcal{F}_{g}{(N)})},{\bf 1})+\mathcal{H}{(\mathcal{F}_{d}{(\mathcal{F}_{g}{(N)})},{\bf 1})},
    \label{eq:hng_train}
\end{equation}
where ${\bf 1}$ indicates that labels associated with the generated negative examples are red tide for $\mathcal{F}_{rtd}$ or real negatives for $\mathcal{F}_{d}$. The generator can be trained by minimizing $\mathcal{L}_{g}(N)$, where $\mathcal{F}_{g}{(N)}$ becomes adversarial examples for the red tide detection model and the discriminator.\medskip

\begin{figure}[t]
\captionsetup{font=small}
\centering
\begin{minipage}[b]{\linewidth}
  \centerline{\includegraphics[width=\linewidth,trim=5mm 5mm 5mm 5mm,clip]{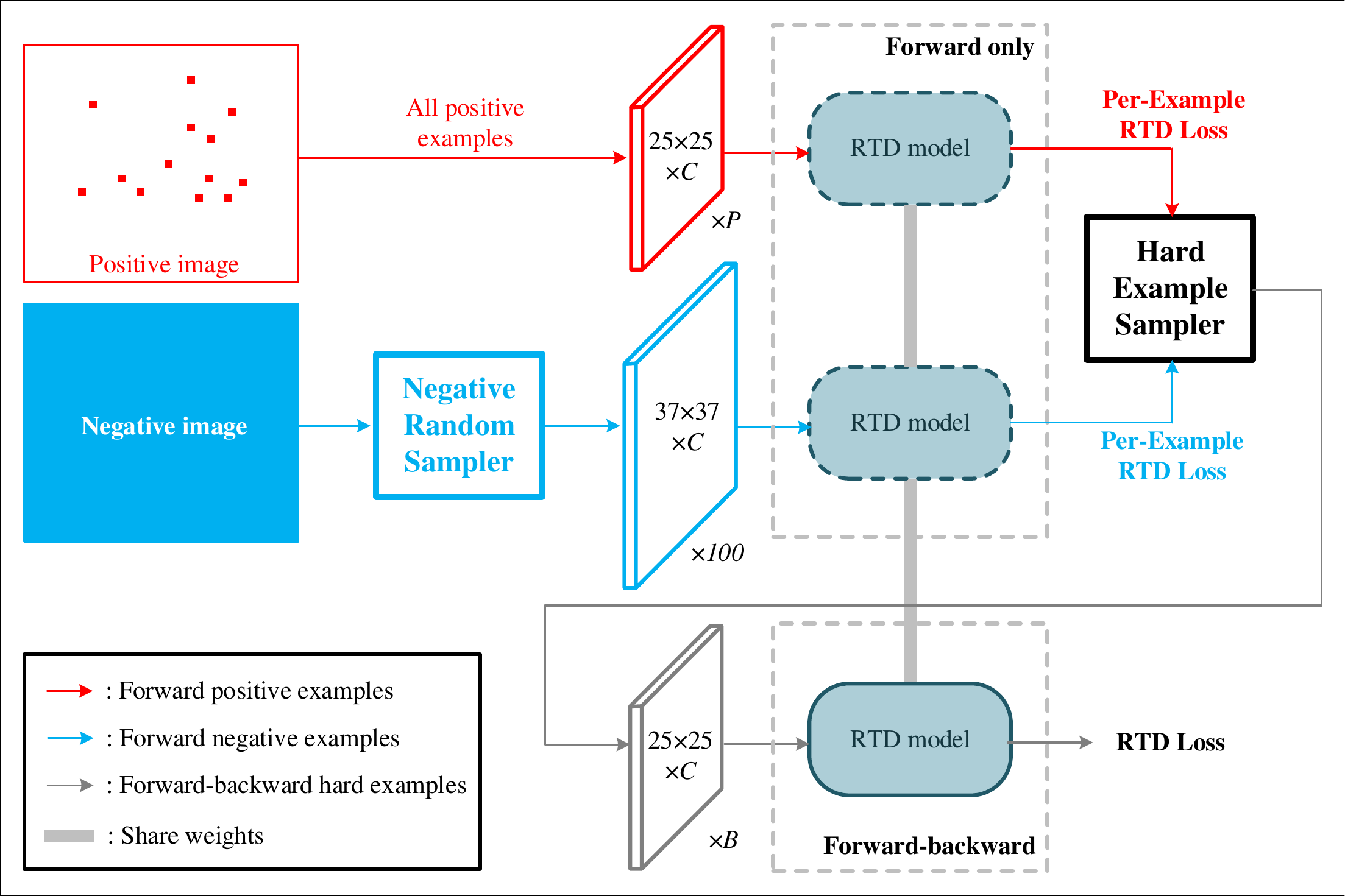}}
\end{minipage}
\caption{{\bf Cascaded online hard example mining} finds hard examples through two-stage sampling (\emph{Negative Random Sampler} and \emph{Hard Example Sampler}) for every iteration. (RTD Model: the proposed red tide detection model, $P$: the size of the positive example set, $B$: batch size, $C$: the spectral dimension of the input example.)}
\label{fig:ohem}
\end{figure}

\noindent{\bf Mining step.} An additional mining step is necessary because the generated examples may not necessarily be hard examples. We use OHEM introduced by Shrivastava et al.~\cite{AShrivastavaCVPR2016} to build batches by collecting hard examples to minimize objective loss throughout the entire examples collectively. However, in problems like ours with many examples, it takes quite long time to find hard examples. In the red tide detection problem, the number of original negative examples in the GOCI images is huge, and even larger-scale generated examples are also considered in training. Therefore, we used OHEM in a cascaded fashion (cOHEM) to first build a pool of randomly chosen negative examples and then carry out OHEM with all the positive examples along with the negative examples in this pool (Figure~\ref{fig:ohem}).

For every pixel labeled with red tide, a 25$\times$25 area around each pixel is collected and used as a positive example. In \emph{Negative Random Sampler}, we randomly select 100 regions of a size 37$\times$37 from a negative image, equivalent to 16,900 examples of size 25$\times$25. In our experiments, we found that it is crucial to randomly select multiple regions instead of one wide region for negative examples, to improve the accuracy of red tide detection using cOHEM effectively. The red tide detection losses for all selected examples are calculated by feeding the examples to the model. Note that this loss represents the extent to which the current model correctly classifies each example. \emph{Hard Example Sampler} randomly selects the high loss examples with predetermined batch size. Then the proposed model is trained with these batches consisting of hard examples.\medskip

\begin{figure}[t]
\captionsetup{font=small}
\centerline{\includegraphics[width=\linewidth,trim=5mm 5mm 5mm 5mm,clip]{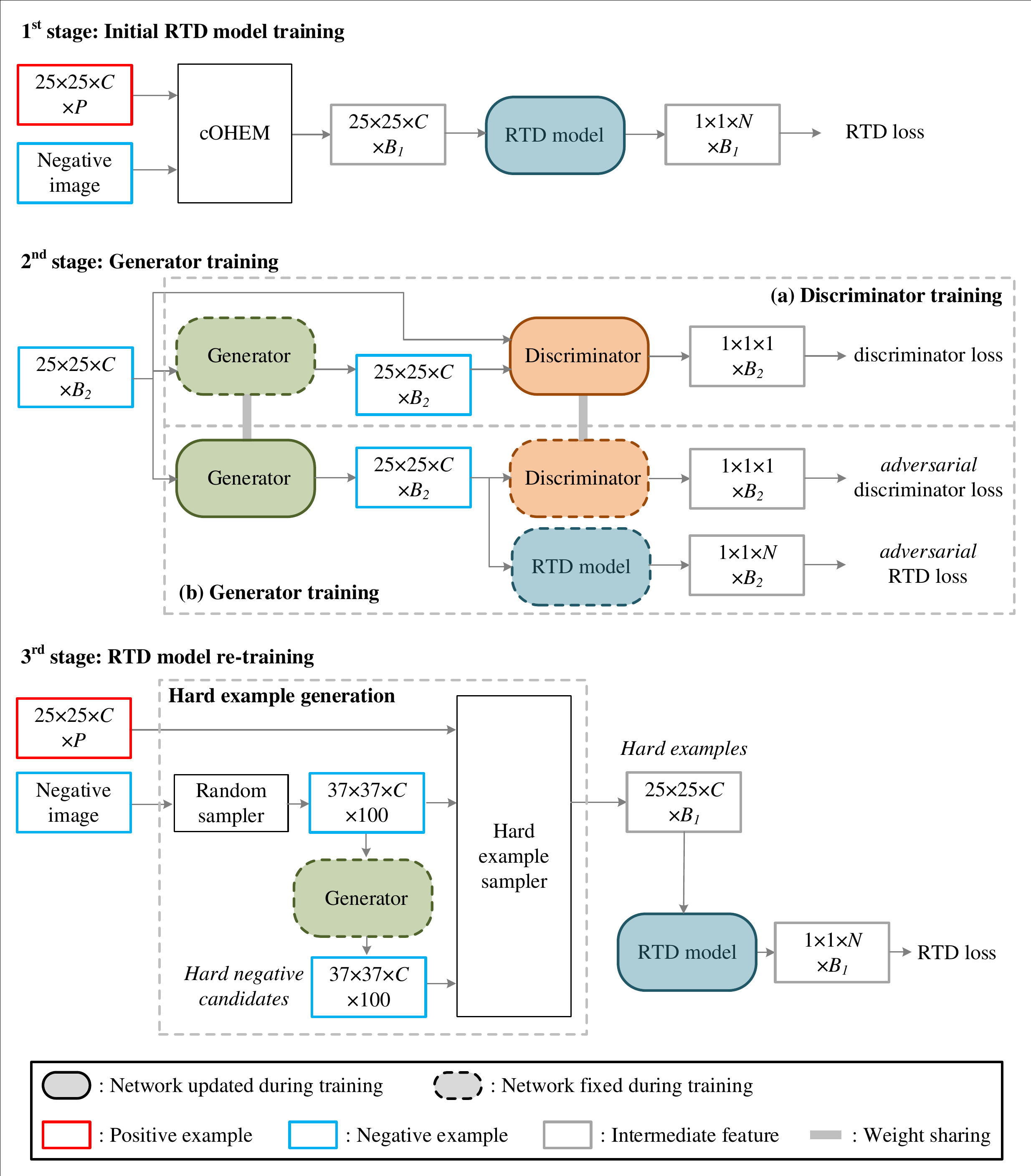}}
\caption{{\bf 3-stage training algorithm.} In the 2nd stage, discriminator training (a) and generator training (b) are performed for each iteration. Data dimension is shown in the data blob in order of height, width, channel, and batch size. $C$, $P$, $N$, $B_1$, and $B_2$ are the spectral dimensions (e.g., 8 for GOCI image), the number of positive examples, the number of categories (e.g., 2 for red tide detection), batch sizes for training RTD model, and batch sizes for training generator, respectively.}
\label{fig:3stage_cascade_training}
\end{figure}

\noindent{\bf 3-stage training strategy.} For training the red tide detection model using the proposed two-step HEG approach, we adopt a 3-stage training strategy. The first stage is to train the initial red tide detection model using cOHEM. In the second stage, the generator and the discriminator are trained, as shown in Figure~\ref{fig:hng}. In this stage, the discriminator is first trained with generator weights unchanged and then the generator is trained while keeping the discriminator and the red tide detection model fixed. In the last stage, the red tide detection model is updated by using hard examples that are the output of the proposed hard example generation approach. Hard examples are mined from real positives, real negatives, and artificially generated negatives via cOHEM. In the third stage, generator weights are fixed.

For the first and third stage, all the 25$\times$25 positive regions and 100 randomly sampled examples of the 37$\times$37 negative regions are used for training. For the second stage, 256 of the 25$\times$25 negative regions are used. Note that our FCN can intake an input image of arbitrary size. The flowchart of the 3-stage training algorithm is shown in Figure~\ref{fig:3stage_cascade_training}. The intermediate output size of the proposed model in each training stage and test are shown in Appendix~\ref{sec:intermediate_output} to understand our proposed model and training strategy better.

\section{Experiments: Red Tide Detection}
\label{sec:experiments}

\subsection{GOCI Satellite Images}
\label{ssec:goci_data}

\begin{figure}[t]
\captionsetup{font=small}
\begin{minipage}[t]{1.0\linewidth}
  \centering
  \centerline{\includegraphics[width=\linewidth,trim=5mm 5mm 5mm 5mm,clip]{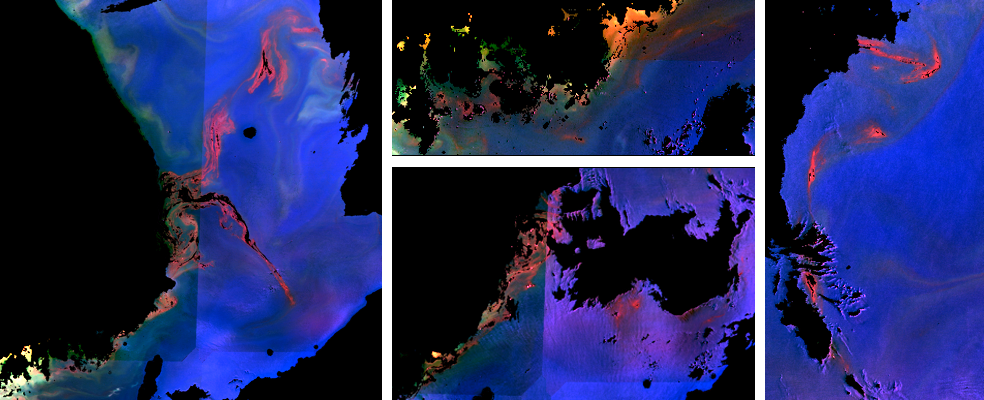}}
\end{minipage}
\caption{{\bf Red Tide Examples Shown on GOCI Images.} In the above figure, the red tide appears as elongated red bands. The images are false color images by combining the 6th, 4th and 1st band of the GOCI multi-spectral image representing the red, green, and blue colors, respectively.}
\label{fig:red_tide}
\end{figure}

GOCI (Geostationary Ocean Color Imager) acquires multispectral images from a large area surrounding the Korean peninsula. The GOCI image~\cite{SChoKJRS2010} has 8 channels consisting of six visible and two near infrared (NIR) frequency bands\footnote{While band extension~\cite{CKwanRS2000,CKwanRS2020,CKwanRS2020b} can be considered because of the small number of bands in the GOCI image, the 9-layer CNN shows sufficient performance even without band extension.} and 500 m spatial resolution. The size of the GOCI image is 5567$\times$5685. Some examples of GOCI images are shown in Figure~\ref{fig:goci_img}. Several red tide examples on GOCI multi-spectral images are also shown in Figure~\ref{fig:red_tide}. 

In this paper, we use GOCI images taken in July, August, and December of 2013 to evaluate our red tide detection model. Images from July and August where red tide occurred are used as positive images, and images from December are used as negative images. Based on some conditions such as the atmosphere, we chose eight images in July and August and four images in December. Half of them were used for training and the other half for testing.

To label red tide pixels, we used the red tide information reported by NIFS (National Institute of Fisheries) of South Korea which directly tested seawater from a ship. NIFS examined red tide occurrence only at a limited number of locations; so it is impossible to cover the entire red tide areas. Furthermore, the red tide positions indicated in the reports were not very accurate due to the error-prone manual process that included mapping geo-coordinates of red tide locations onto GOCI images. Hence, we have extended potential red tide regions up to 25 km (50 pixel distance) in all directions from the red tide location indicated in the report and then labeled red tide with experts' help. Approximately 100 pixels from each training image were sparsely labeled as a red tide area. We used pixels labeled as red tide as positive examples and all the pixels of the December images as negative examples.

\begin{table*}[!t]
\captionsetup{font=small}
\caption{{\bf The Optimum Specification of the Red Tide Detection Model.} The model specification is empirically determined based on three measures: detection accuracy (AUC), training time, and test time. Titan XP is used as a GPU that affects training and test time. {\bf Numbers in bold} indicate the specifications used in our model. $\sim$13$\times$13 means the multi-scale filter bank consisting of 1$\times$1, 5$\times$5, 9$\times$9, and 13$\times$13 convolutional filters.}
\begin{centering}
\begin{tabular}{ccc}
\subfloat[\# of filters of each layer]{
\setlength{\tabcolsep}{4.5pt}
\renewcommand{\arraystretch}{1.2}
\begin{tabular}{r|cccc}
& 64 & 96 & {\bf 128} & 192 \\\specialrule{.15em}{.05em}{.05em} 
AUC (\%) & 79.6 & 83.5 & {\bf 87.1} & 87.0 \\
Train (min) & 42.0 & 49.1 & {\bf 56.4} & 62.5\\
Test (sec/im) & 13.4 & 16.2 & {\bf 18.3} & 20.5 \\
\end{tabular}
}&
\subfloat[\# of residual module]{
\setlength{\tabcolsep}{4.5pt}
\renewcommand{\arraystretch}{1.2}
\begin{tabular}{r|ccc}
& 1 & {\bf 2} & 3 \\\specialrule{.15em}{.05em}{.05em} 
AUC (\%) & 81.0 & {\bf 87.1} & 86.6 \\
Train (min) & 47.5 & {\bf 56.4} & 63.1 \\
Test (sec/im) & 16.0 & {\bf 18.3} & 21.4 \\
\end{tabular}
}&
\subfloat[Multi-scale filter bank.]{
\setlength{\tabcolsep}{4.5pt}
\renewcommand{\arraystretch}{1.2}
\begin{tabular}{r|cccc}
& 1$\times$1 & $\sim$5$\times$5 & $\sim$9$\times$9 & {\bf $\sim$13$\times$13} \\\specialrule{.15em}{.05em}{.05em} 
AUC (\%) & 85.0 & 85.3 & 86.7 & {\bf 87.1} \\
Train (min) & 37.5 & 40.5 & 47.5 & {\bf 56.4} \\
Test (sec/im) & 11.3 & 12.5 & 14.3 & {\bf 18.3} \\
\end{tabular}
}
\end{tabular}
\end{centering}
\label{tab:network_spec}
\end{table*}

\subsection{Evaluation Settings}
\label{ssec:setting}

\noindent{\bf Evaluation metrics.} We used two different metrics to evaluate the proposed model: the receiver operating characteristic (ROC) curve and the ROC variation curve. The ROC variation curve describes changes in the detection rate based on varying numbers of (true or false) detections per image (NDPI) instead of the false positive rate. This metric is beneficial when there are numerous unlabeled examples whose identity is unknown. Note that in a GOCI image only a fraction of red tide pixels are labeled and the rest of the image remains unlabeled. For quantitative analysis, we calculate the AUC (the area under the ROC curve) and ndpi@dr=0.25, ndpi@dr=0.5 and ndpi@dr=0.75 indicating the NDPI values when the detection rate reaches 0.25, 0.5, and 0.75, respectively.\medskip

\noindent{\bf Model training.} The proposed models are trained from scratch. When HEG is used, we used a three-stage training strategy and trained the model with 1250 iterations for each stage. A base learning rate is 0.01 for the red tide detection model and generator and 0.0001 for the discriminator. The base learning rate drops to a factor of 10 for every 500 iterations. When the three stage training strategy is not used to train the model (i.e. artificially generated examples are not used for training), we trained the model with 2500 iterations. A base learning rate is 0.01 and drops by a factor of 10 for every 1K iterations.

The proposed models are optimized by using a mini-batch Stochastic Gradient Descent (SGD) approach with a batch size of 256 examples, the momentum of 0.9, and weight decay of 0.0005. The red tide detection model's training objective is to minimize the cross entropy losses between the red tide labels and the final output scores. Each batch consists of examples extracted from one positive image with red tide occurrence and one negative image with no red tide occurrence. The positive-to-negative ratio in each batch is set to 1:3.

To reduce overfitting in training, data augmentation is carried out. Since a GOCI image is taken from a top view, training examples are augmented by mirroring across the horizontal, vertical, and diagonal axes. This mirroring can be performed in one direction or in multiple directions. This will increase the number of examples by eight times.

When training the red tide detection model, all learnable layers except for the layers of residual modules ($3^{rd}$, $4^{th}$, $5^{th}$, and $6^{th}$ layers) are initialized according to Gaussian distribution with zero mean and 0.01 standard deviation. The layers of the residual modules are initialized according to Gaussian distribution with a mean of zero and a standard deviation of 0.005. All layers of the generator except for the last layer are initialized according to Gaussian distribution with a mean of zero and a standard deviation of 0.02. The last layer is initialized according to Gaussian distribution with a mean of zero and a standard deviation of 50.

\subsection{Architecture Design}
\label{ssec:architecture_design}

In this section, we use a 2-fold cross validation that splits the training set into two subsets, alternating one for training and another for testing. AUC reported in the tables is the average over two validations.\medskip

\begin{table*}[t]
\captionsetup{font=small}
\setlength{\tabcolsep}{18.0pt}
\renewcommand{\arraystretch}{1.2}
\caption{{\bf The Optimum Specification of the Generator.} For each architecture, the numbers in parentheses indicate the number of filters in the deconvolutional layers. The discriminator architecture is designed by adding one fully connected layer to the first four layers of the generator.}
\vspace{-0.3cm}
\begin{center}
\begin{tabular}{c|c|c}
Architecture & AUC (\%) & Train (sec/iter)\\
\specialrule{.15em}{.05em}{.05em}
16$\rightarrow$16$\rightarrow$32$\rightarrow$32($\rightarrow$32)$\rightarrow$16$\rightarrow$16($\rightarrow$16)$\rightarrow$8$\rightarrow$8 & 84.3 & 42.5 \\
32$\rightarrow$32$\rightarrow$64$\rightarrow$64($\rightarrow$64)$\rightarrow$32$\rightarrow$32($\rightarrow$32)$\rightarrow$16$\rightarrow$8 & 88.6 & 47.6 \\
{\bf 64$\rightarrow$64$\rightarrow$128$\rightarrow$128($\rightarrow$128)$\rightarrow$64$\rightarrow$64($\rightarrow$64)$\rightarrow$32$\rightarrow$8} & {\bf 92.0} & {\bf 51.8} \\
128$\rightarrow$128$\rightarrow$256$\rightarrow$256($\rightarrow$256)$\rightarrow$128$\rightarrow$128($\rightarrow$128)$\rightarrow$64$\rightarrow$8 & 91.5 & 55.3 \\
\end{tabular}
\end{center}
\label{tab:HNG_spec}
\end{table*}

\noindent{\bf Finding the model specification.} To find the optimal specification of the red tide detection model, we evaluate the model by changing various types of model parameters, such as the number of filters and residual modules, and the types of filters used in the multi-scale filter bank. The proposed model specifications are determined by evaluating detection accuracy (AUC), training time, and test time on various model parameters. The final model specification used in the proposed work is shown in Table~\ref{tab:network_spec}. Table~\ref{tab:network_spec} also indicates if a more extensive network is used by increasing the depth and breadth of the network, network overfitting on the GOCI dataset starts to occur. We use one GPU (NVIDIA Titan XP) that affects training and test times to conduct our experiments.

We also optimize the generator by changing the number of filters. As shown in Figure \ref{fig:hng}, the generator is designed as a conv-deconv network consisting of eight convolutional layers and two deconvolutional layers. In this architecture, the number of filters in the first layer $n$ is doubled in the third layer and then reduced by half in the sixth and again in the ninth layer. The last layer has eight filters so that its output has the same eight channels as those of the GOCI image's spectral signal. We evaluate detection accuracy and training time to find optimal architecture while $n$ is varied among 16, 32, 64, and 128, as shown in Table~\ref{tab:HNG_spec}. Based on the results in Table~\ref{tab:HNG_spec}, we used the third architecture that adopts 64, 64, 128, 128, 128, 64, 64, 64, 32, and 8 filters for all ten layers, respectively. The fourth architecture, which employs the largest number of filters, may overfit the GOCI data set.\medskip

\begin{table}[t]
\captionsetup{font=small}
\setlength{\tabcolsep}{8.5pt}
\renewcommand{\arraystretch}{1.2}
\caption{{\bf Accuracy of Various Negative Random Sampling Strategies of cOHEM.} {\bf Numbers in bold} indicate our cOHEM sampling strategy.}
\vspace{-0.2cm}
\begin{center}
\begin{tabular}{r|ccc|c}
Window size & 153$\times$153 & 65$\times$65 & {\bf 37$\times$37} & 25$\times$25 \\
\# of window & 1 & 10 & {\bf 100} & 192 \\\specialrule{.15em}{.05em}{.05em}
\# of pixels & 22.9K & 41.3K & {\bf 133.7K} & 120.0K \\
\# of examples & 16.3K & 16.4K & {\bf 16.5K} & 192 \\\hline
AUC (\%) & 87.7 & 89.1 & {\bf 90.4} & 88.5 \\
Train time (sec/iter) & 0.521 & 0.534 & {\bf 0.540} & 0.482 \\
\end{tabular}
\end{center}
\label{tab:cOHEM_strategy}
\end{table}

\noindent{\bf Finding sampling strategy of mining.} We compare various negative example sampling strategies of cOHEM with respect to detection accuracy and training time. Negative example sampling can change based on two factors: window size and the number of windows. In Table~\ref{tab:cOHEM_strategy}, we compare four different sampling strategies with various factors. The first three strategies are chosen to maintain a similar number of negative examples as used in training. The last strategy (192 windows of 25$\times$25) is training without using cOHEM. Accordingly, its training time is the shortest.

In Table~\ref{tab:cOHEM_strategy}, the third strategy gives the best performance in terms of AUC and training time. This observation indicates that increasing spatial diversity of sampling is essential in providing competitive performance. Therefore, even though the third strategy requires a large amount of memory due to the large pixels, it is adopted in our training approach.

\begin{table*}
\captionsetup{font=small}
\begin{minipage}{\linewidth}
\setlength{\tabcolsep}{14pt}
\renewcommand{\arraystretch}{1.2}
\caption{{\bf Red Tide Detection Accuracy.} For each metric, {\bf numbers in bold} indicate the best accuracy. Note that the higher the AUC value, the better the performance, and the smaller the value of NDPI, the better the performance.}
\centering
\begin{tabular}{l|cccc}
\multicolumn{1}{c|}{Method} & AUC (\%) & ndpi@dr=0.25 & ndpi@dr=0.5 & ndpi@dr=0.75 \\
\specialrule{.15em}{.05em}{.05em}
SVM & 81.0 & 295732 & 4641404 & 31648395 \\
~~~~~~+ Hard Negative Mining & 82.4 & 114067 & 1259805 & 5313930 \\\hline
\cite{HLeeTIP2017} & 84.9 & 38642 & 165402 & 623806 \\
~~~~~~+ cOHEM & 87.5 & 27320 & 121269 & 356460 \\
~~~~~~+ HEG & 88.9 & 16990 & 67792 & 261054 \\\hline
DR-CNN\cite{MZhangTIP2018} & 89.4 & 15381 & 62933 & 254869 \\
~~~~~~+ cOHEM & 91.1 & 12057 & 49507 & 129733 \\
~~~~~~+ HEG & 92.9 & 5890 & 33696 & 149099 \\\hline
Ours & 90.6 & 8307 & 33080 & 131789 \\
~~~~~~+ cOHEM & 93.2 & 8298 & 30917 & 77196 \\
~~~~~~+ HEG & {\bf 95.0} & {\bf 5722} & {\bf 13168} & {\bf 50220} \\
\end{tabular}
\label{tab:performance}
\end{minipage}
\end{table*}

\subsection{Experimental Results}
\label{ssec:performance}

\noindent{\bf Baselines.} We implement three baselines: SVM and two CNN-based hyperspectral image classification approach~\cite{HLeeTIP2017,MZhangTIP2018}. In SVM, a 25$\times$25 region centered on the pixel in test is used as a feature representing the center pixel. To know the advantages of adopting hard example mining, we applied conventional hard negative mining~\cite{KSungMITAIMemo1994} to SVM training. \cite{HLeeTIP2017} is the CNN-based approach by which our model has been inspired. Another CNN-based baseline, Diverse Region-based CNN (DR-CNN)~\cite{MZhangTIP2018}, inputs a set of diverse regions consisting of six different regions (i.e., global , right, left, top, bottom, and local regions) to encode semantic context-aware representation. In this experiment, we use a 25$\times$25 image patch as a global region compatible with the input dimensions of our approach. 13$\times$25 sub-patch at top and bottom of the patch are used as the top and bottom regions. Similarly, 25$\times$13 sub-patch at the left and right of the global patch are the left and right regions, respectively. The 3$\times$3 region at the center of the global patch is used as a local region.\medskip

\noindent{\bf Performance comparison.} Table~\ref{tab:performance} shows that our model trained using hard examples via HEG provides the highest accuracy in all four metrics. The proposed HEG was effective in improving the performance of our model and two CNN-based baselines. It is also observed that adopting a hard example mining approach consistently improves the accuracy of all four methods as it efficiently eases the significant imbalance between red tide and non-red tide examples. 

\begin{figure}[t]
\captionsetup{font=small}
\begin{minipage}{\linewidth}
\centering
\includegraphics[width=\linewidth,trim=22mm 85mm 30mm 85mm,clip]{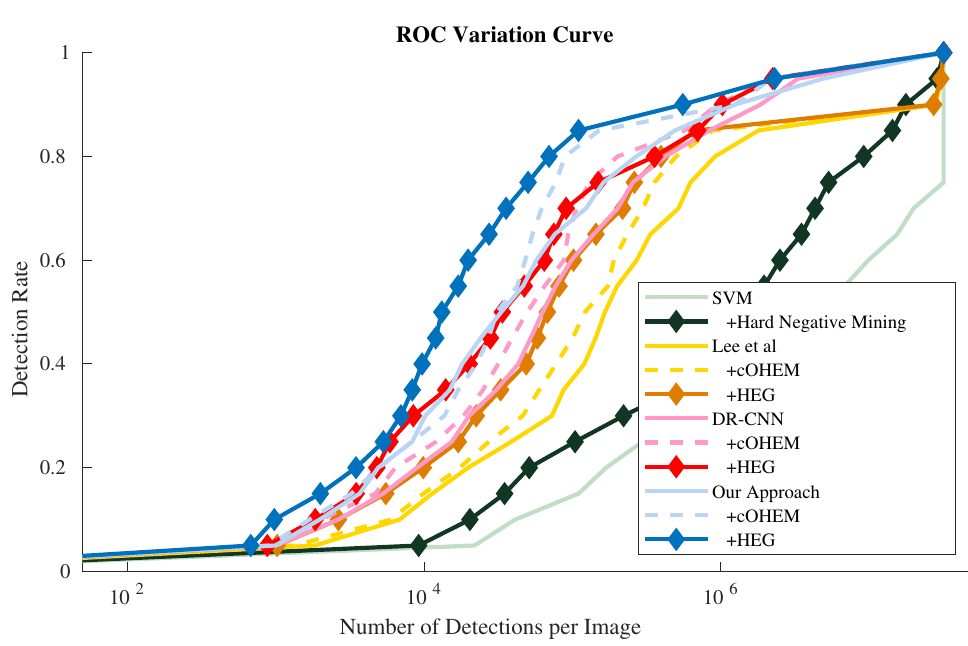}
\caption{{\bf ROC Variation Curve.} To better view changes in detection rate and NDPI in detail, the $x$ axis is shown with a logarithmic scale.}
\label{fig:roc}
\end{minipage}
\end{figure}

\begin{figure}[t]
\captionsetup{font=small}
  \centering
\begin{minipage}[b]{\linewidth}
  \centerline{\includegraphics[width=\linewidth,trim=0mm 5mm 0mm 5mm,clip]{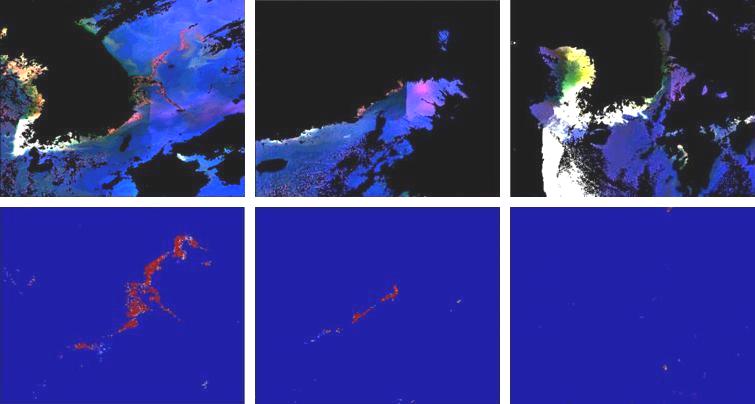}}
\end{minipage}
\caption{{\bf Red Tide Detection Results.} The top images and the bottom images are input images and the corresponding red tide detection results from our approach, respectively. The first and second input images have red tide in them and the last image is a negative image without red tide.}
\label{fig:results}
\end{figure}

Figure~\ref{fig:roc} shows the ROC variation curves for our model and baselines. From Figure~\ref{fig:roc}, we can confirm that our model provides significantly enhanced detection performance compared to the baselines over the most range of NDPI. Figure~\ref{fig:results} shows red tide detection results from our approach.

\subsection{Analyzing Hard Negative Candidates}

\begin{figure}[t]
\captionsetup{font=small}
\begin{minipage}[b]{\linewidth}
  \centerline{\includegraphics[width=\linewidth,trim=55mm 114mm 62mm 114mm,clip]{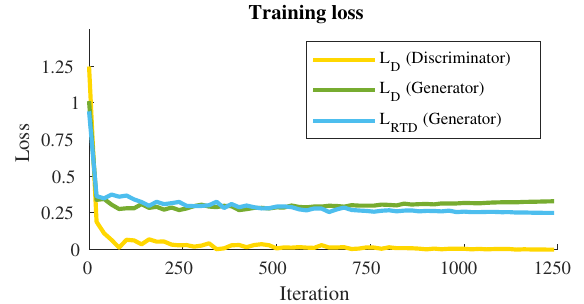}}
\end{minipage}
\caption{{\bf Training Curve.} Losses changed during training generator and discriminator in the 2nd training stage are shown.}
\label{fig:training_loss}
\end{figure}

In Figure~\ref{fig:training_loss}, we observe that the generator and discriminator converge successfully in the $2^{nd}$ training stage. This shows that generator optimization overcomes the interference of the red tide detection model and the discriminator.

\begin{figure*}[t]
\captionsetup{font=small}
\begin{minipage}{\linewidth}
\centerline{\includegraphics[width=\linewidth,trim=0mm 5mm 0mm 5mm,clip]{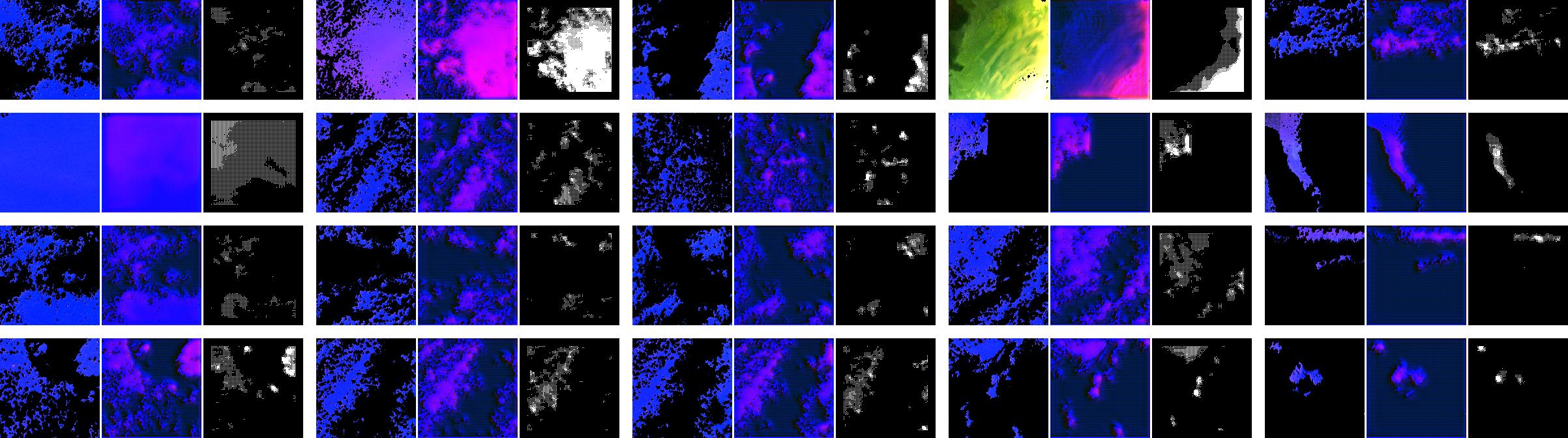}}
\caption{{\bf Generated Training Examples (after 2nd Training Stage).} Each set consists of three images: real negative example (left), generated negative example (center), and red tide detection model activation by the generated example (right). White pixels in the activation map indicate pixels with a red tide score estimated by the red tide detection model greater than 0.5. For every set in this figure, there was no activation for a real negative example.}
\label{fig:gen_examples}
\end{minipage}
\end{figure*}

\begin{figure}[t]
\captionsetup{font=small}
\begin{minipage}{\linewidth}
\centerline{\includegraphics[width=\linewidth,trim=35mm 100mm 42mm 100mm,clip]{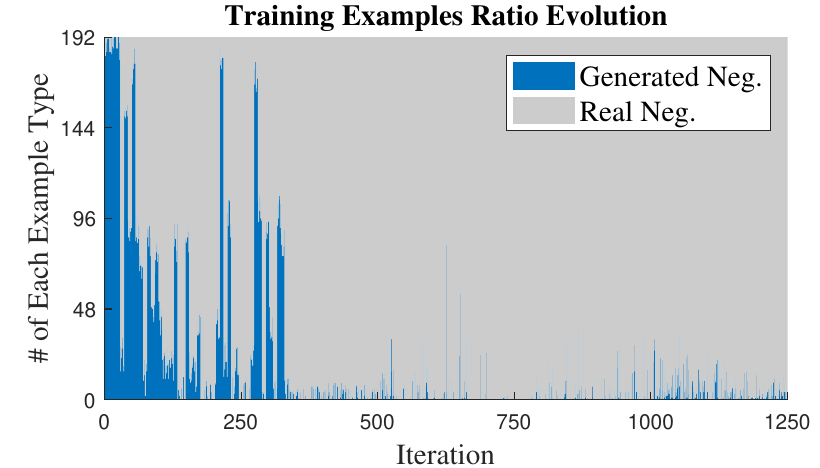}}
\caption{{\bf Negative Example Ratio Evolution (in 3rd Training Stage).} This evolution shows the change in training negative example ratio at the third stage of the training strategy.}
\label{fig:training_example_portion}
\end{minipage}
\end{figure}

\begin{figure*}[t!]
\captionsetup{font=small}
  \centering
  \includegraphics[width=0.119\linewidth,trim=70mm 100mm 67mm 92mm,clip]{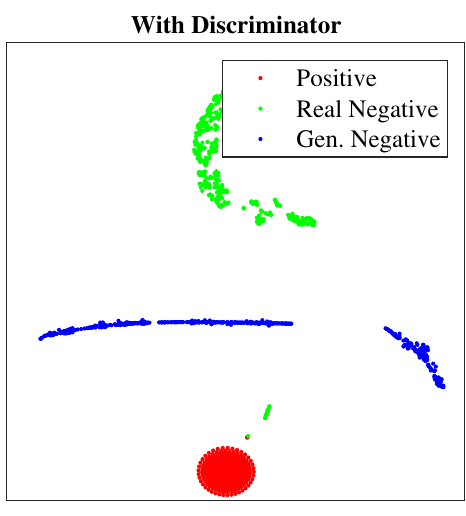}~
  \includegraphics[width=0.119\linewidth,trim=70mm 100mm 67mm 92mm,clip]{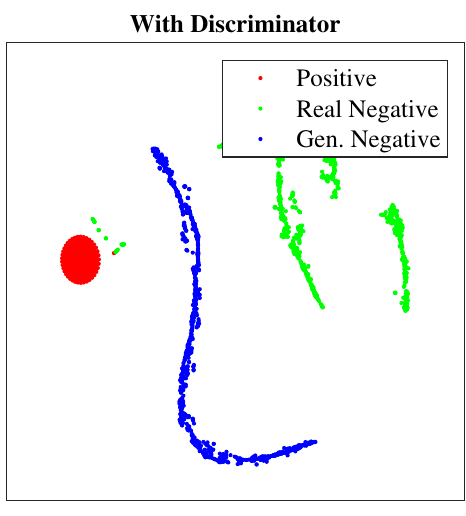}~
  \includegraphics[width=0.119\linewidth,trim=70mm 100mm 67mm 92mm,clip]{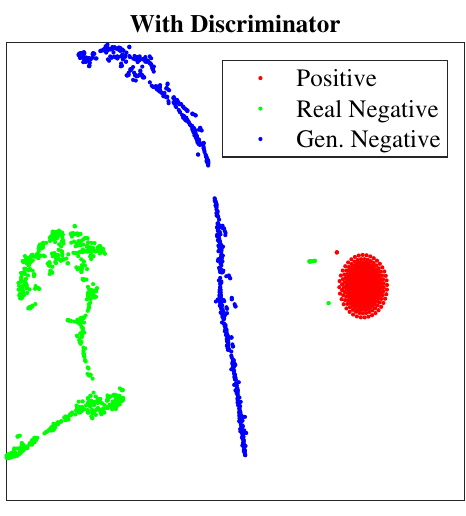}~
  \includegraphics[width=0.119\linewidth,trim=70mm 100mm 67mm 92mm,clip]{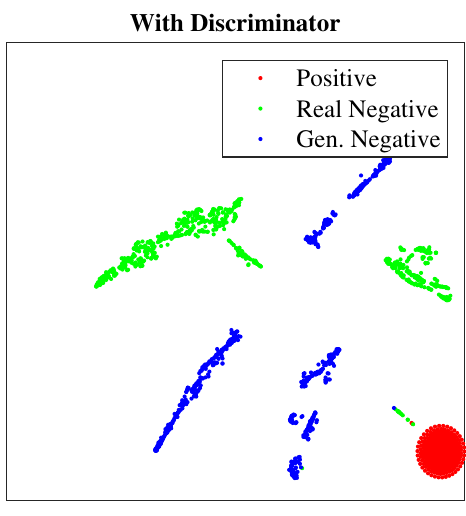}~
  \includegraphics[width=0.119\linewidth,trim=70mm 100mm 67mm 92mm,clip]{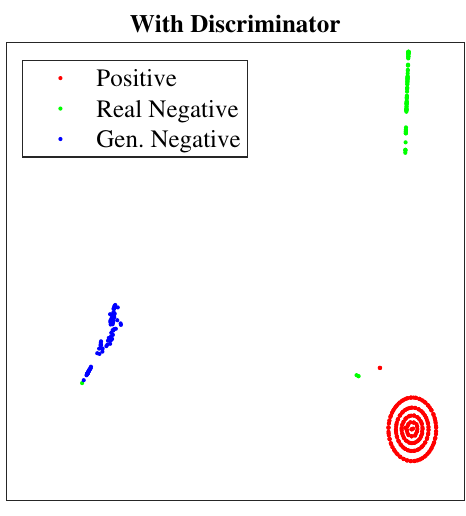}~
  \includegraphics[width=0.119\linewidth,trim=70mm 100mm 67mm 92mm,clip]{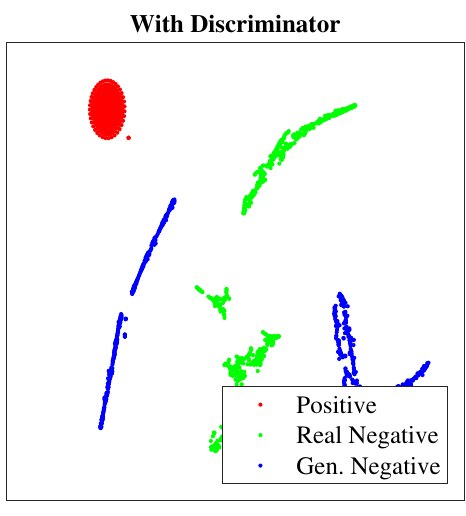}~
  \includegraphics[width=0.119\linewidth,trim=70mm 100mm 67mm 92mm,clip]{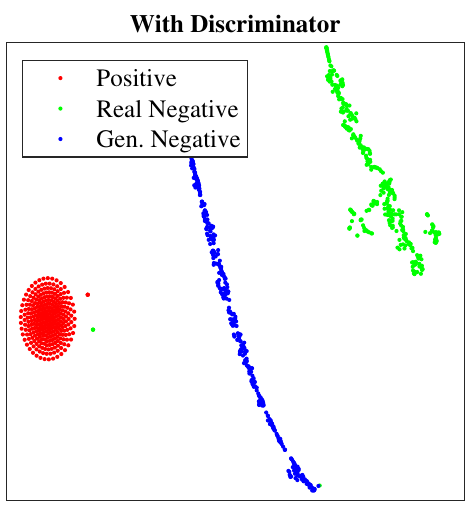}~
  \includegraphics[width=0.119\linewidth,trim=70mm 100mm 67mm 92mm,clip]{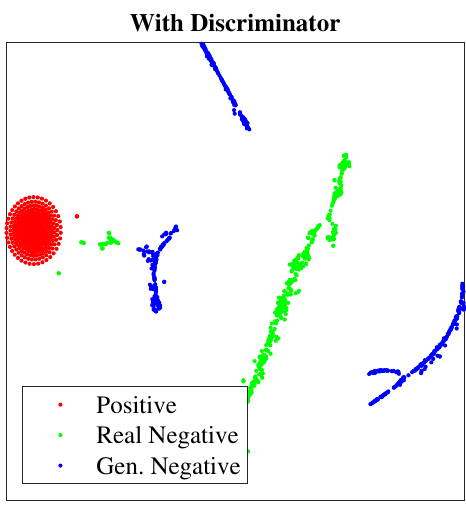}\\
  \includegraphics[width=0.119\linewidth,trim=70mm 100mm 67mm 92mm,clip]{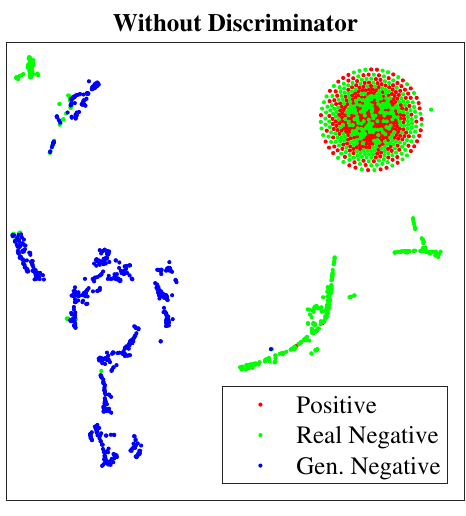}~
  \includegraphics[width=0.119\linewidth,trim=70mm 100mm 67mm 92mm,clip]{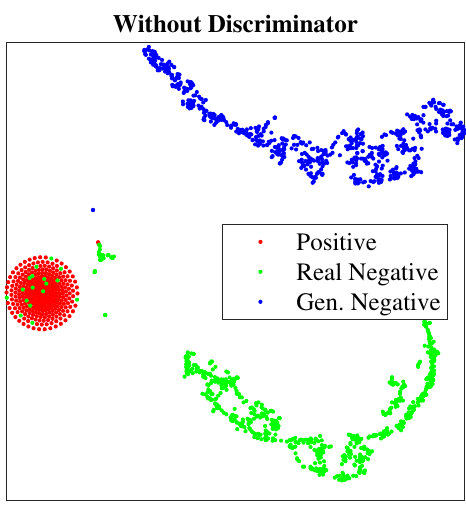}~
  \includegraphics[width=0.119\linewidth,trim=70mm 100mm 67mm 92mm,clip]{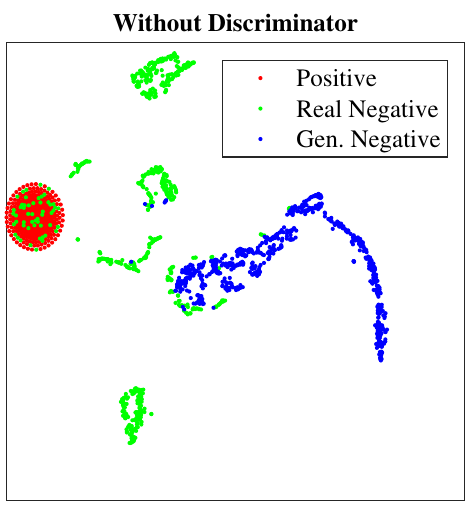}~
  \includegraphics[width=0.119\linewidth,trim=70mm 100mm 67mm 92mm,clip]{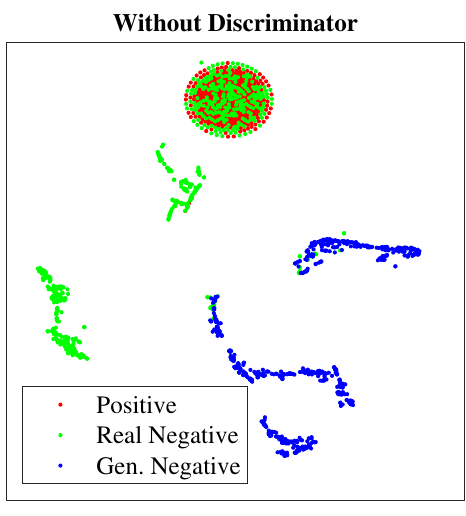}~
  \includegraphics[width=0.119\linewidth,trim=70mm 100mm 67mm 92mm,clip]{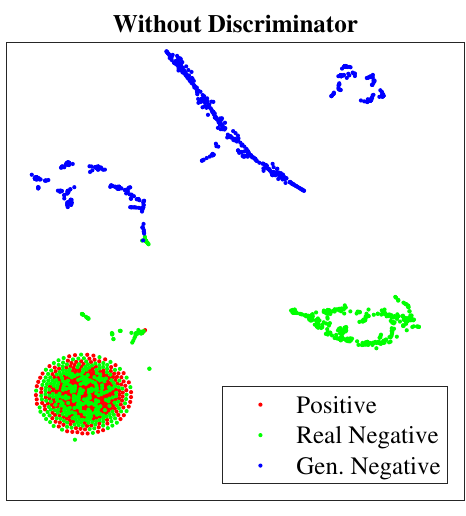}~
  \includegraphics[width=0.119\linewidth,trim=70mm 100mm 67mm 92mm,clip]{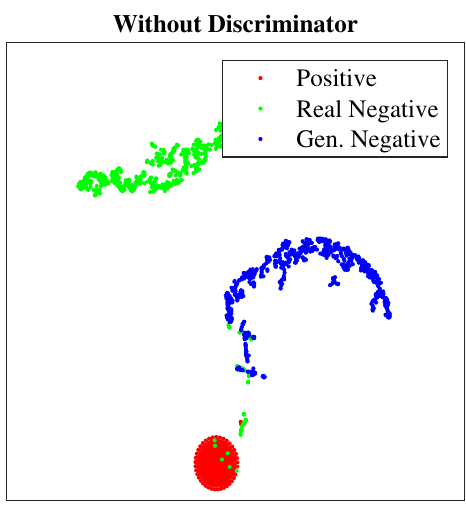}~
  \includegraphics[width=0.119\linewidth,trim=70mm 100mm 67mm 92mm,clip]{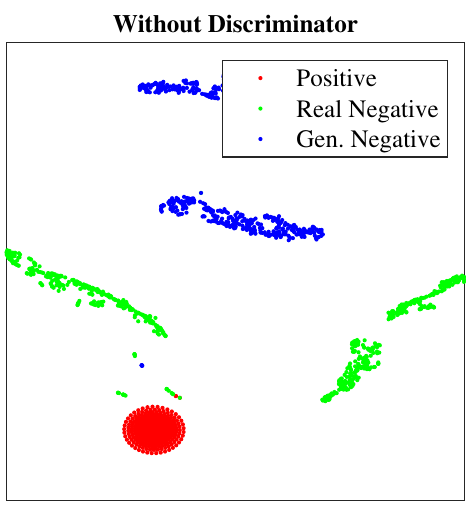}~
  \includegraphics[width=0.119\linewidth,trim=70mm 100mm 67mm 92mm,clip]{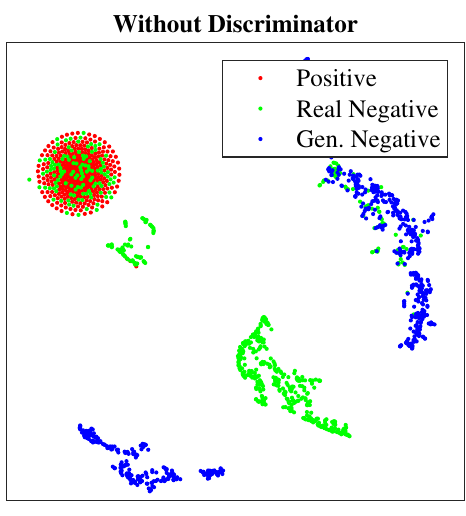}~
\caption{{\bf t-SNE Visualization of Three Different Types of Training Examples.} The top row cases indicate when the discriminator was used for training the generator, and the bottom row cases show when the discriminator was not used. Data points close to each other in the embedded space are likely to have similar spectral characteristics.}
\label{fig:tsne_analysis}
\end{figure*}

Some examples of generated hard negative candidates, the successfully trained generator's output, are presented in Figure~\ref{fig:gen_examples}. In Figure~\ref{fig:gen_examples}, it can be observed that activated pixel regions (the $3^{rd}$ column of each set) tend to have different colors--i.e., different intensity values for some spectral bands--for certain areas than the ones covered by real non-red tide examples. Note that the activated regions in Figure~\ref{fig:gen_examples} do not show typical characteristics of red tide regions--narrow elongated bands with sharp boundaries as shown in Figure~\ref{fig:red_tide}. Furthermore, the activated regions by artificial non-red examples appear to be pink while the real red tide is generally red.

Figure~\ref{fig:training_example_portion} shows an evolution of the ratio of training examples between real negative examples and generated negative examples within one batch as the $3^{rd}$ training stage progresses. Overall, the proportion of generated negative examples within a batch gradually decreases as the training iteration increases. In Figure~\ref{fig:training_example_portion}, the generated examples appear to be heavily used for optimization in the first few iterations. After about 350 iterations, the positive examples are successfully separated from the generated examples by the optimized network, providing higher accuracy than the case in which the generated examples are not used for training.

To analyze the relationship among three different types of training examples--i.e. positive, real negative, and artificially generated negative examples-- after training the generator at the 2nd training stage, we use a visualization technique called t-distributed stochastic neighborhood embedding (t-SNE)~\cite{LMaatenJMLR2008}, as shown in Figure~\ref{fig:tsne_analysis}. For the visualization, 500 examples are randomly chosen separately from real positive and negative examples. The artificially generated examples are the output of the generator, which takes real negative examples as input. This selection of only a small fraction of real negative examples--i.e., about 500 from 400M examples-- is due to the significant imbalance between the number of positive and negative examples. To address this issue, we provide multiple instances of the tSNE-based visualization using different sets of randomly selected training examples to illustrate the general relationship, as shown in Figure~\ref{fig:tsne_analysis}. We use the output of the $8^{th}$ convolutional layer (after ReLU layer) as the selected examples' features. As shown on the top rows of Figure~\ref{fig:tsne_analysis}, the artificially generated negative examples are placed between the positive and real negative examples. This indicates that the generated negative examples can be considered hard example candidates when retraining the red tide detection model. When training the generator without using the discriminator, the artificially generated negative examples are no longer hard examples as they are placed far to the right of the positive and real negative examples, as shown on the bottom rows of Figure~\ref{fig:tsne_analysis}.

\subsection{Ablation Study}
\label{ssec:neg_of_pos_img}

\noindent{\bf Training with unlabeled examples.} The main problem with using GOCI images is the labeling of red tide pixels. It is quite challenging to label every pixel where a red tide appears on GOCI due to practical issues. Therefore, there is no guaranty that pixels not labeled as the red tide in positive images (i.e. red tide images) are non-red tide pixels.

We carried out ablation experiments to validate our claim that pixels not labeled as red tide in positive images should not be used to train the proposed model. When unlabeled pixels are used, we also used all pixels of negative images for training. Table~\ref{tab:neg_of_pos_img} provides accuracy with and without the use of unlabeled examples for training with regard to the four evaluation metrics. Table~\ref{tab:neg_of_pos_img} shows that using unlabeled examples significantly reduces the accuracy for all the evaluation metrics. The unlabeled examples including red tide pixels, adversely affect the proposed model' training, supporting the reason why unlabeled pixels should not be used for training.\medskip

\begin{table}[t]
\captionsetup{font=small}
\setlength{\tabcolsep}{8.0pt}
\renewcommand{\arraystretch}{1.2}
\caption{{\bf Accuracy with and without Unlabeled Examples.} {\bf Numbers in bold} indicate the best accuracy for each evaluation metric.}
\vspace{-0.3cm}
\begin{center}
\begin{tabular}{r|cccc}
Training Examples & AUC & \begin{tabular}{@{}c@{}}npdi \\ @dr=0.25\end{tabular} & \begin{tabular}{@{}c@{}}npdi \\ @dr=0.5\end{tabular} & \begin{tabular}{@{}c@{}}npdi \\ @dr=0.75\end{tabular}\\\specialrule{.15em}{.05em}{.05em}
w/ Unlabeled & 82.8 & 105472 & 952403 & 1608326 \\
w/o Unlabeled & \bf{90.6} & \bf{8307} & \bf{33080} & \bf{131789} \\
\end{tabular}
\end{center}
\label{tab:neg_of_pos_img}
\end{table}

\begin{table}[t]
\captionsetup{font=small}
\setlength{\tabcolsep}{10.5pt}
\renewcommand{\arraystretch}{1.2}
\caption{{\bf Comparison with GAN-based negative generator.} w/o HEG indicates the case where generated negative examples are not used for training. w/ GAN is the case using generated examples via GAN where the generator is trained only to fool the discriminator. w/ Ours uses hard negative examples generated by our approach. {\bf Numbers in bold} indicate the best accuracy for each evaluation metric.}
\vspace{-0.3cm}
\begin{center}
\begin{tabular}{c|cccc}
HEG & AUC & \begin{tabular}{@{}c@{}}npdi \\ @dr=0.25\end{tabular} & \begin{tabular}{@{}c@{}}npdi \\ @dr=0.5\end{tabular} & \begin{tabular}{@{}c@{}}npdi \\ @dr=0.75\end{tabular} \\\specialrule{.15em}{.05em}{.05em}
w/o HEG & 90.6 & 8307 & 33080 & 131789 \\
w/ GAN & 90.3 & 8797 & 36217 & 129536 \\
w/ Ours & \bf{95.0} & \bf{5722} & \bf{13168} & \bf{50220} \\
\end{tabular}
\end{center}
\label{tab:comparison_with_GAN}
\end{table}

\noindent{\bf Generating negative examples with the discriminator only.} To demonstrate the effectiveness of fooling red tide detector for generating hard negative example candidates, we also test a generator trained by fooling the discriminator only. The generator training approach becomes GAN~\cite{IGoodfellowNIPS2014} in that it consists of a discriminator and a generator, and these networks learn to restrain each other. Augmenting the training data by GAN-based image generation increased accuracy in many approaches~\cite{ADosovitskiyNIPS2016,ANguyenCVPR2017,LHHughesRS2018,ADibaCVPR2019}. However, Table~\ref{tab:comparison_with_GAN} shows that the GAN-based data augmentation approach is significantly less performed than our approach. Moreover, it was even worse than the case without using generated examples. These observations indicate that deceiving the red tide detector in training the generator is essential for improving accuracy.

\begin{figure}[t]
\captionsetup{font=small}
  \centering
\begin{minipage}[b]{\linewidth}
  \centerline{\includegraphics[width=\linewidth,trim=5mm 5mm 5mm 5mm,clip]{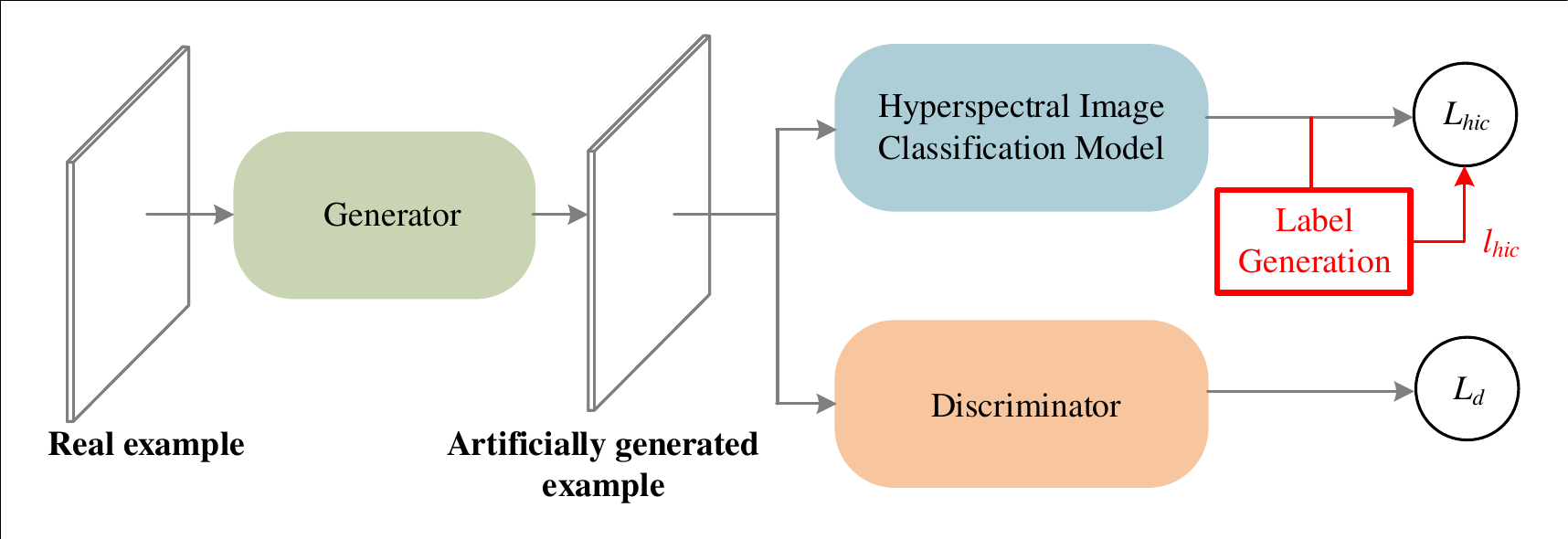}}
\end{minipage}
\caption{{\bf Training multi-category generator} follows the same process as hard negative candidate generator training shown in Figure~\ref{fig:hng}. The only difference is to add the ``Label Generation'' module, which calculates the adversarial label for each example, at the end of the hyperspectral image classification model.}
\label{fig:heg}
\end{figure}

\begin{table*}[t!]
\captionsetup{font=small}
\caption{{\bf Selected Classes for Evaluation and the Numbers of Training and Test Examples.}}
\begin{tabular}{ccc}
\subfloat[Indian Pines]{
\setlength{\tabcolsep}{7.0pt}
\renewcommand{\arraystretch}{1.2}
\begin{tabular}[t]{lcc}
Class & Training & Test \\\specialrule{.15em}{.05em}{.05em} 
Corn-notill & 200 & 1228 \\
Corn-mintill & 200 & 630 \\
Grass-pasture & 200 & 283 \\
Hay-windrowed & 200 & 278 \\
Soybean-notill & 200 & 772 \\
Soybean-mintill & 200 & 2255 \\
Soybean-clean & 200 & 393 \\
Woods & 200 & 1065 \\\specialrule{.15em}{.05em}{.05em} 
Total & 1600 & 6904 \\
\end{tabular}
}
&
\subfloat[Salinas]{
\setlength{\tabcolsep}{7.0pt}
\renewcommand{\arraystretch}{1.2}
\begin{tabular}[t]{lcc}
Class & Training & Test \\\specialrule{.15em}{.05em}{.05em} 
Brocooli green weeds 1 & 200 & 1809 \\
Brocooli green weeds 2 & 200 & 3526 \\
Fallow & 200 & 1776 \\
Fallow rough plow & 200 & 1194 \\
Fallow smooth & 200 & 2478 \\
Stubble & 200 & 3759 \\
Celery & 200 & 3379 \\
Grapes untrained & 200 & 11071 \\
Soil vineyard develop & 200 & 6003 \\
Corn senesced green weeds & 200 & 3078 \\
Lettuce romaines, 4 wk & 200 & 868 \\
Lettuce romaines, 5 wk & 200 & 1727 \\
Lettuce romaines, 6 wk & 200 & 716 \\
Lettuce romaines, 7 wk & 200 & 870 \\
Vineyard untrained & 200 & 7068 \\
Vineyard vertical trellis & 200 & 1607 \\\specialrule{.15em}{.05em}{.05em} 
Total & 3200 & 50929 \\
\end{tabular}
}
&
\subfloat[PaviaU]{
\setlength{\tabcolsep}{7.0pt}
\renewcommand{\arraystretch}{1.2}
\begin{tabular}[t]{lcc}
Class & Training & Test \\\specialrule{.15em}{.05em}{.05em} 
Asphalt & 200 & 6431 \\
Meadows & 200 & 18449 \\
Gravel & 200 & 1899 \\
Trees & 200 & 2864 \\
Sheets & 200 & 1145 \\
Bare soils & 200 & 4829 \\
Bitumen & 200 & 1130 \\
Bricks & 200 & 2482 \\
Shadows & 200 & 747 \\\specialrule{.15em}{.05em}{.05em} 
Total & 1800 & 40976 \\
\end{tabular}
}
\end{tabular}
\label{tab:dataset_info}
\end{table*}

\section{Experiments: Hyperspectral Image Classification}
\label{sec:hyperspectral}

Our approach can also be easily generalized to hyperspectral image (HSI) classification, requiring pixel-wise classification. For the HSI classification, we have used three benchmark datasets: Indian pines, Salinas, and PaviaU. For each HSI dataset, 200 pixels randomly selected from each category are used for training and all the remaining pixels are used for testing.\medskip

\noindent{\bf Evaluation setting.} For hyperspectral image classification, we carried out experiments on three hyperspectral datasets: Indian Pines, Salinas, and PaviaU. The Indian Pines dataset is an image consisting of 145$\times$145 pixels with 200 spectral reflectance bands covering the spectral range from 0.4 to 2.5 $\mu$m with a spatial resolution of 20 m. There are 16 material categories in the Indian Pines dataset, but only eight materials with relatively large samples are used for evaluation. The Salinas dataset includes 16 classes with 512$\times$217 pixels, 204 spectral bands, and a high spatial resolution of 3.7 m. The Salinas and the Indian Pines datasets have the same frequency characteristics because they are acquired by the same AVIRIS sensor. The PaviaU dataset acquired by ROSIS sensor has 610$\times$340 pixels with nine material categories and 103 spectral bands covering 0.43 to 0.86 $\mu$m spectral range with a 1.3 m spatial resolution. For the Salinas dataset and the PaviaU dataset, unlike the Indian Pines dataset, all available material categories are used.

For each dataset, randomly selected 200 examples from each category are used for training and the remaining examples are used for test, as shown in Table~\ref{tab:dataset_info}. We perform this partitioning 20 times and present the mean and standard deviation as overall classification accuracy.\medskip

\noindent{\bf Hard example candidate generation.} When applying the proposed training strategy to hyperspectral image classification, we replace single-category hard negative generator with a multi-category hard example candidate generator that generates hard examples of individual material categories instead of hard negatives. The generator is trained to create examples misclassified by the hyperspectral image classification model. For each example $e$, its adversarial training label ($l^a_{hic}$) is the category with the largest loss among all categories except for the class the example belongs to:
\begin{equation}
l^{a}_{hic} = \argmax_{l\in C,~l\neq l_{hic}}{\mathcal{L}_{hic}(e,l)},\label{eq:heg}
\end{equation}
where $l_{hic}$ is a label associated with the example $e$ and $C$ is the set of all categories. Generator optimization is carried out with Equation~\ref{eq:hng_train} by replacing $L_{rtd}$ (denoted as {\bf 1} in the Equation) with $L^{a}_{hic}=\{l^{a}_{hic}\}$.\medskip

\noindent{\bf Architecture.} Unlike the red tide detection model that outputs one-dimensional sigmoid probabilities, the hyperspectral image classification model uses the softmax layer to calculate multiple categories' final probabilities. We also use the softmax loss to train the model while the red tide detection model is optimized by minimizing the cross-entropy loss.

When training the generator in the 2nd training stage, we add the module (``Label Generation`` in Figure~\ref{fig:heg}) to the end of the network. This module finds an adversarial label for each example using Equation \ref{eq:heg}. The example becomes a hard example for its adversarial label when training the model. While the generator used in red tide detection only intakes negative examples to generate hard negative examples, the generator for hyperspectral image classification takes examples of any material category as an input.\medskip

\noindent{\bf Model training.} A stochastic gradient descent (SGD) approach is used to train the model for hyperspectral image classification. For each iteration in training, a batch includes 256 examples. The base learning rate is set to 0.001 and reduced by a factor of 10 for every step size. For the other optimization parameters, we set the momentum to 0.9, gamma to 0.1, and weight decay to 0.0005. Table~\ref{tab:Iteration} shows the step size and the number of iterations for each training stage.\medskip

\noindent{\bf Results.} Table \ref{tab:HICarrucary} shows the classification accuracy for the three datasets. For all the three datasets, our 9-layer CNN outperforms the baseline introduced in \cite{HLeeTIP2017}. Our strategy of generating hard example generation further improves performance by at least 0.44\%.

Note that the problems we encountered with GOCI images (i.e. extreme sparsity of red tide samples, significant imbalanced distribution, difficulties in accurate groundtruthing, etc.) are not normally observed in other hyperspectral images. Enhanced performance for HSI classification verifies that the proposed hard example generation approach is also applicable to other related problem domains, such as HSI analysis.

\section{Conclusion}
\label{sec:conclusion}

In this paper, we have developed a novel 9-layer fully convolutional network (FCN) suitable for pixel-wise classification. Due to the challenges of annotating Earth-observing remotely sensed images to the pixel level, there are very few fully annotated satellite-based remote sensing data. To avoid the performance degradation caused by significantly insufficient and imbalanced training data, we introduce a novel approach based on hard example generation (HEG). The proposed HEG approach takes two steps, first generating hard example candidates and mining hard examples from real and generated examples. In the first step, the generator that creates hard example candidates is learned via the adversarial learning framework by fooling a discriminator and a pixel-wise classification model at the same time. In the second step, online hard example mining is used in a cascaded fashion to mine hard examples from a large pool of real and artificially generated examples. The proposed FCN jointly trained with HEG approach provides state-of-the-art accuracy for red tide detection. We also show that the proposed approach can be easily extended to other tasks, such as hyperspectral image classification.

\begin{table}[t!]
\captionsetup{font=small}
\setlength{\tabcolsep}{5.2pt}
\renewcommand{\arraystretch}{1.2}
\caption{{\bf The Step Size and the Number of Iteration} for different training sets.}
\begin{tabular}{c|cc|cc|cc}
Dataset & \multicolumn{2}{c|}{Indian Pines} & \multicolumn{2}{c|}{Salinas} & \multicolumn{2}{c}{PaviaU} \\
Stage & 1 \& 3 & 2 & 1 \& 3 & 2 & 1 \& 3 & 2 \\\specialrule{.15em}{.05em}{.05em} 
SS/Iter. & 500/1250 & 500/1250 & 2K/5K & 800/3K & 2K/5K & 800/3K \\
\end{tabular}
\label{tab:Iteration}
\end{table}

\begin{table}[t]
\captionsetup{font=small}
\setlength{\tabcolsep}{9.5pt}
\renewcommand{\arraystretch}{1.2}
\caption{{\bf Hyperspectral Image Classification Accuracy (Mean and Standard Deviation).} When calculating the accuracy for our model, training and evaluation protocols follow those in \cite{HLeeTIP2017}.}
\vspace{-0.3cm}
\begin{center}
\begin{tabular}{c|c|cc}
Dataset & \cite{HLeeTIP2017} & Ours & +HEG\\
\specialrule{.15em}{.05em}{.05em}
Indian pines & 93.61 $\pm$ 0.56 &  95.17 $\pm$ 0.48 & \bf{96.43} $\pm$ \bf{0.42} \\
Salinas & 95.07 $\pm$ 0.23 & 96.09 $\pm$ 0.40 & \bf{96.53} $\pm$ \bf{0.48}\\
PaviaU & 95.97 $\pm$ 0.46 & 96.27 $\pm$ 0.51 & \bf{97.33} $\pm$ \bf{0.37}\\
\end{tabular}
\end{center}
\label{tab:HICarrucary}
\end{table}

\begin{table*}[t]
\captionsetup{font=small}
\setlength{\tabcolsep}{0.79pt}
\renewcommand{\arraystretch}{1.2}
\caption{{\bf The Size of the Intermediate Output.} The numbers in parentheses indicate the width, height, and channels of the intermediate output, respectively. $C$, $D$, and $FC$ denote the convolutional layer, the deconvolutional layer, and the fully connected layer, respectively.}
\scriptsize{
\begin{tabular}{c|rl|c|cc|cc|cc|c|c|c|c}
\multicolumn{14}{c}{{\bf Training: Stage 1 \& 3 (Training Red Tide Detection Network by Using cOHEM)}}\\\specialrule{.15em}{.05em}{.05em} 
\multicolumn{14}{c}{Before Hard Example Sampling}\\\specialrule{.15em}{.05em}{.05em} 
 & \multicolumn{2}{c|}{Input} & $C$1$\times$1 & $C$5$\times$5 (pad: 4) & MAX5$\times$5 & $C$9$\times$9 (pad: 8) & MAX9$\times$9 & $C$13$\times$13 (pad: 12) & MAX13$\times$13 & Concat. & $C$2,$\cdots$,$C$8 & $C$9 & Output \\\specialrule{.15em}{.05em}{.05em} 
Positive & \multicolumn{2}{c|}{(25,25,8)} & (25,25,128) & (29,29,128) & (25,25,128) & (33,33,128) & (25,25,128) & (37,37,128) & (25,25,128) & (25,25,512) & (25,25,128) & (25,25,1) & (1,1,1) \\\hline
Negative & \multicolumn{2}{c|}{(37,37,8)} & (37,37,128) & (41,41,128) & (37,37,128) & (45,45,128) & (37,37,128) & (49,49,128) & (37,37,128) & (37,37,512) & (37,37,128) & (37,37,1) & (13,13,1) \\\specialrule{.15em}{.05em}{.05em} 
\multicolumn{14}{c}{After Hard Example Sampling}\\\specialrule{.15em}{.05em}{.05em} 
 & \multicolumn{2}{c|}{Input} & $C$1$\times$1 & $C$5$\times$5 & MAX5$\times$5 & $C$9$\times$9 & MAX9$\times$9 & $C$13$\times$13 & MAX13$\times$13 & Concat. & $C$2,$\cdots$,$C$8 & $C$9 & Output \\\specialrule{.15em}{.05em}{.05em} 
\multirow{4}{*}{Hard Example} & \multirow{4}{*}{(25,25,8)} & (1,1,8) & (1,1,8) & $\cdot$ & $\cdot$ & $\cdot$ & $\cdot$ & $\cdot$ & $\cdot$ & \multirow{4}{*}{(1,1,512)} & \multirow{4}{*}{(1,1,128)} & \multirow{4}{*}{(1,1,1)} & \multirow{4}{*}{(1,1,1)} \\
 & & (9,9,8) & $\cdot$ & (5,5,128) & (1,1,128) & $\cdot$ & $\cdot$ & $\cdot$ & $\cdot$ & & & & \\
 & & (17,17,8) & $\cdot$ & $\cdot$ & $\cdot$ & (9,9,128) & (1,1,128) & $\cdot$ & $\cdot$ & & & & \\
 & & (25,25,8) & $\cdot$ & $\cdot$ & $\cdot$ & $\cdot$ & $\cdot$ & (13,13,128) & (1,1,128) & & & & \\
\specialrule{.15em}{.05em}{.05em}
\end{tabular}

\setlength{\tabcolsep}{1.85pt}
\begin{tabular}{c|c|c|cc|ccc|c|c|c|c|c|c|rl|c}
\multicolumn{17}{c}{}\\
\multicolumn{17}{c}{{\bf Training: Stage 2 (Training HNG/Discriminator)}} \\\specialrule{.15em}{.05em}{.05em}
\multicolumn{9}{c|}{HNG} & \multicolumn{5}{c|}{Discriminator} & \multicolumn{3}{c}{RT Detect Net.}\\\hline
Input & $C$1, $C$2 & $C$3, $C$4 & $D$5 & $C$6, $C$7 & $D$8 & $C$9 & $C$10 & Output & Input & $C$1, $C$2 & $C$3, $C$4 & $FC$ & Output & \multicolumn{2}{c|}{Input} & $C$1$\times$1 $\cdots$ \\\specialrule{.15em}{.05em}{.05em} 
\multirow{4}{*}{(25,25,8)} & \multirow{4}{*}{(13,13,64)} & \multirow{4}{*}{(7,7,128)} & \multirow{4}{*}{(13,13,128)} & \multirow{4}{*}{(13,13,64)} & \multirow{4}{*}{(25,25,64)} & \multirow{4}{*}{(25,25,32)} & \multirow{4}{*}{(25,25,8)} & \multirow{4}{*}{(25,25,8)} & \multirow{4}{*}{(25,25,8)} & \multirow{4}{*}{(13,13,64)} & \multirow{4}{*}{(7,7,128)} & \multirow{4}{*}{(1)} & \multirow{4}{*}{(1)} & \multirow{4}{*}{(25,25,8)} & (1,1,8) & \multirow{4}{*}{$\cdots$} \\
& & & & & & & & & & & & & & & (9,9,8) & \\
& & & & & & & & & & & & & & & (17,17,8) & \\
& & & & & & & & & & & & & & & (25,25,8) & \\
\specialrule{.15em}{.05em}{.05em}
\end{tabular}

\setlength{\tabcolsep}{14.0pt}
\begin{tabular}{c|c|c|c|c|c|c|c|c}
\multicolumn{9}{c}{}\\
\multicolumn{9}{c}{{\bf Training: Stage 3 (Artificially Generating Hard Examples)}} \\\specialrule{.15em}{.05em}{.05em}
\multicolumn{9}{c}{HNG} \\\hline
Input & $C$1, $C$2 & $C$3, $C$4 & $D$5 & $C$6, $C$7 & $D$8 & $C$9 & $C$10 & Output\\\specialrule{.15em}{.05em}{.05em} 
(37,37,8) & (19,19,64) & (10,10,128) & (19,19,128) & (19,19,64) & (37,37,64) & (37,37,32) & (37,37,8) & (37,37,8) \\
\specialrule{.15em}{.05em}{.05em}
\end{tabular}

\setlength{\tabcolsep}{1.3pt}
\begin{tabular}{c|c|cc|cc|cc|c|c|c|c}
\multicolumn{12}{c}{}\\
\multicolumn{12}{c}{{\bf Test}}\\\specialrule{.15em}{.05em}{.05em}
Input & $C$1$\times$1 & $C$5$\times$5 (pad: 4) & MAX5$\times$5 & $C$9$\times$9 (pad: 8) & MAX9$\times$9 & $C$13$\times$13 (pad: 12) & MAX13$\times$13 & Concat. & $C$2,$\cdots$,$C$8 & $C$9 & Output \\\specialrule{.15em}{.05em}{.05em} 
(600,600,8) & (600,600,128) & (604,604,128) & (600,600,128) & (608,608,128) & (600,600,128) & (612,612,128) & (600,600,128) & (600,600,512) & (600,600,128) & (600,600,1) & (576,576,1) \\
\specialrule{.15em}{.05em}{.05em} 
\end{tabular}
}
\label{tab:output_size}
\end{table*}

\appendix

\section{Intermediate Output Size of the Proposed Models}
\label{sec:intermediate_output}

Our FCN architecture is designed to be slightly different between training and test to make it suitable for pixel-wise classification. The size of the input is also different for individual training stages. Furthermore, in the first and third training stages, the sizes of positive and negative examples are different for the same stages, making it easy to understand the architecture. Accordingly, to get an accurate understanding of the architecture, we provide the size of the intermediate output of the model in Table~\ref{tab:output_size}. The red tide detection model weights are transferred between training and testing and during different training phases.

\ifCLASSOPTIONcompsoc
  \section*{Acknowledgments}
\else
  \section*{Acknowledgment}
\fi

This research was supported by the ``Development of the integrated data processing system for GOCI-II'' funded by the Ministry of Ocean and Fisheries, Korea.


%

\ifCLASSOPTIONcaptionsoff
  \newpage
\fi



\bibliographystyle{IEEEtran}
\bibliography{egbib.bib}
\end{document}